\documentclass[sigconf]{acmart}
\renewcommand\footnotetextcopyrightpermission[1]{}
\AtBeginDocument{%
  }

\newcommand{\xmark}{\texttimes}
\usepackage{makecell}  
\usepackage{float}
\usepackage{array} 
\usepackage{booktabs}
\usepackage{multirow}
\usepackage{subcaption}
\usepackage[table]{xcolor}
\usepackage{algorithm}
\usepackage{algpseudocode}
\usepackage{listings}
\usepackage{xcolor}
\usepackage{placeins}

\setcopyright{acmlicensed}
\copyrightyear{2018}
\acmYear{2018}
\acmDOI{XXXXXXX.XXXXXXX}
\acmConference[Conference acronym 'XX]{Make sure to enter the correct
  conference title from your rights confirmation email}{June 03--05,
  2018}{Woodstock, NY}
\acmISBN{978-1-4503-XXXX-X/2018/06}

\begin{document}

\title{MissBench: Benchmarking Multimodal Affective Analysis \\under Imbalanced Missing Modalities}

\author{Tien Anh Pham}
\affiliation{%
  \institution{VNU University of Engineering and Technology}
  \city{Hanoi}
  \country{Vietnam}
}
\email{23021696@vnu.edu.vn}

\author{Phuong-Anh Nguyen}
\affiliation{%
  \institution{VNU University of Engineering and Technology}
  \city{Hanoi}
  \country{Vietnam}
}
\email{22028332@vnu.edu.vn}

\author{Duc-Trong Le}
\affiliation{%
  \institution{VNU University of Engineering and Technology}
  \city{Hanoi}
  \country{Vietnam}
}
\email{trongld@vnu.edu.vn}

\author{Cam-Van Thi Nguyen}
\authornote{Corresponding author.}
\affiliation{%
  \institution{VNU University of Engineering and Technology}
  \city{Hanoi}
  \country{Vietnam}
}
\email{vanntc@vnu.edu.vn}

\renewcommand{\shortauthors}{Pham et al.}

\begin{abstract}
Multimodal affective computing underpins key tasks such as sentiment analysis and emotion recognition. Standard evaluations, however, often assume that textual, acoustic, and visual modalities are equally available. In real applications, some modalities are systematically more fragile or expensive, creating imbalanced missing rates and training biases that task-level metrics alone do not reveal.
We introduce MissBench, a benchmark and framework for multimodal affective tasks that standardizes both shared and imbalanced missing-rate protocols on four widely used sentiment and emotion datasets. MissBench also defines two diagnostic metrics. The Modality Equity Index (MEI) measures how fairly different modalities contribute across missing-modality configurations. The Modality Learning Index (MLI) quantifies optimization imbalance by comparing modality-specific gradient norms during training, aggregated across modality-related modules. Experiments on representative method families show that models that appear robust under shared missing rates can still exhibit marked modality inequity and optimization imbalance under imbalanced conditions. These findings position MissBench, together with MEI and MLI, as practical tools for stress-testing and analyzing multimodal affective models in realistic incomplete-modality settings.  For reproducibility, we release our code at: \url{https://anonymous.4open.science/r/MissBench-4098/}
\end{abstract}



\maketitle

\section{Introduction}
\begin{figure}[ht!]
    \centering
    \includegraphics[width=\linewidth]{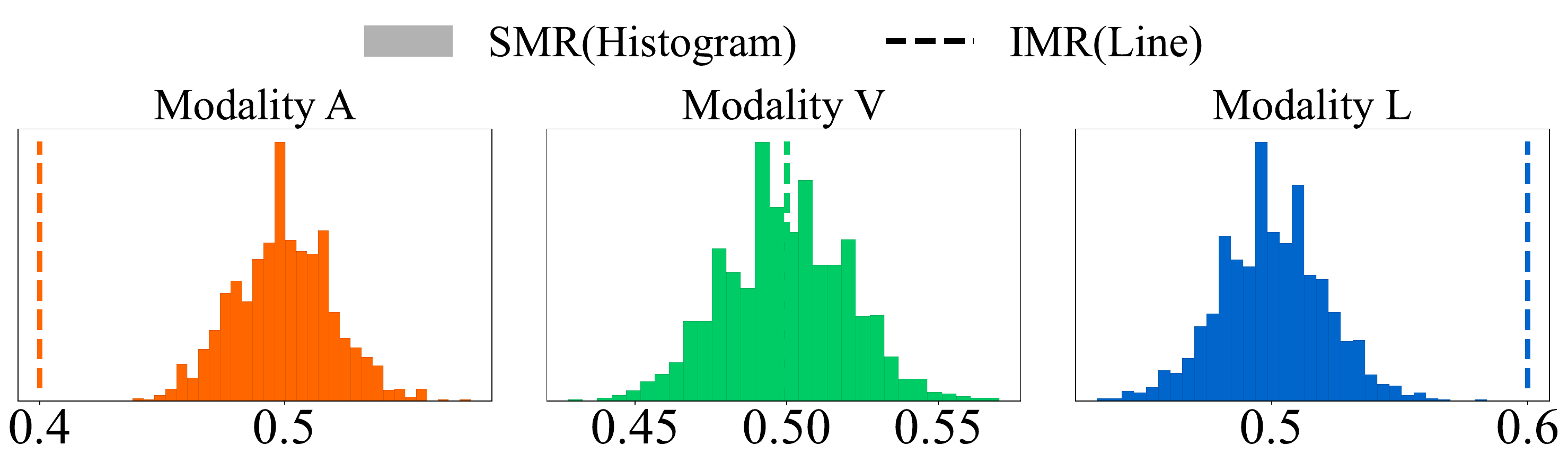}
    \caption{Sampled modality-specific missing rate under mean-match SMR=$0.5$ and IMR=$(0.4,0.5,0.6)$. Under SMR $\hat{r_m}\approx r_{\mathrm{sh}}$, while under IMR $r_m$ can receive any predefined value $\in[0,1)$.}
    \label{fig:motiv-ratio}
\end{figure}

\begin{figure}[htbp]
    \centering
    \includegraphics[width=\linewidth]{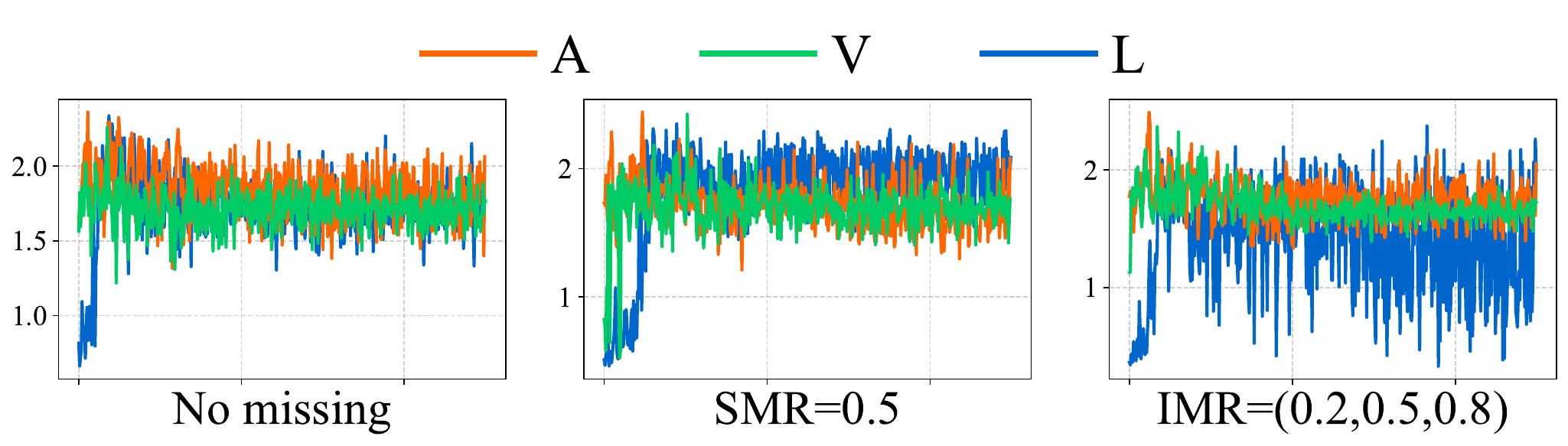}
    \caption{Decoupled modality parameters~\cite{wang2023unlocking} of RedCore \cite{sun2024redcore} on IEMOCAP under different missing-rate configurations, illustrating how a single dominant modality can drive most parameter updates under imbalanced missing rates.}
    \label{fig:motiv-grad}
\end{figure}
Multimodal affective computing underpins key applications such as sentiment analysis and emotion recognition in dialog systems, call centers, and social media \cite{hu2024recent, poria2017review, zhao2024deep}. These systems typically integrate language, visual, and acoustic modalities, yet in realistic environments data from some modalities are often missing or severely degraded due to sensor failures, noise, occlusion, or privacy and transcription constraints \cite{wu2024deep,zhang2024multimodal}. As a result, learning from incomplete multimodal inputs is essential for building reliable affective models \cite{liang2024foundations, lin2024adapt, ma2022multimodal}.

A large body of work has been devoted to improving robustness under missing modalities, including imputation strategies, reconstruction-based models, and architectures designed to operate on arbitrary modality subsets \cite{ma2021smil,zhao2021MMIN,lian2023gcnet,reza2024robust}. While these methods substantially improve performance compared to naive fusion, their evaluation predominantly relies on task-level metrics such as Accuracy, F1-score, MAE, or correlation. Such aggregate metrics summarize end performance but offer limited insight into \emph{how} different modalities contribute, or how learning dynamics change when missingness is systematic rather than random \cite{wu2024deep, wu2022characterizing, ma2022multimodal}.

\begin{table*}[t!]
\centering
\caption{Comparison with prior work on missing modalities, multimodal imbalance, and benchmarks.
``Missing protocol'' shows whether SMR or IMR is explicitly modeled.
``Mod.-level'' and ``Grad.-level'' indicate modality-wise and optimization-level diagnostics comparable to odality Equity Index (MEI) and Modality Learning Index (MLI).}
\resizebox{0.85\linewidth}{!}{%
\begin{tabular}{lcccccc}
\toprule
Work & Affective & Missing & IMR & Mod.-level & Grad.-level & Public \\
 & focus & protocol & (imb.) & diagnostics & diagnostics & benchmark \\
\midrule
MMIN~\cite{zhao2021MMIN} & \checkmark & SMR / pattern & \xmark & \xmark & \xmark & \xmark \\
GCNet~\cite{lian2023gcnet} & \checkmark & SMR / fixed & \xmark & \xmark & \xmark & \xmark \\
RedCore~\cite{sun2024redcore} & \checkmark & IMR (custom) & \checkmark & limited & \xmark & \xmark \\
MCE~\cite{zhao2025mce} & general & IMR (general) & \checkmark & utility metric & gradient-aware & \xmark \\
MultiBench~\cite{liang2021multibench} & mixed & noise / drop & \xmark & \xmark & \xmark & \checkmark \\
MERBench~\cite{lian2026merbench} & \checkmark & noise / robustness & \xmark & \xmark & \xmark & \checkmark \\
MissMAC-Bench~\cite{lin2026missmac} 
& \checkmark 
& fixed \& random MR 
& \xmark 
& perf. mean \& std 
& \xmark 
& \checkmark \\
BalanceBenchmark~\cite{xu2025balancebenchmark} & general & imbalance & \xmark & imbalance index & \xmark & \checkmark \\
\midrule
\textbf{MissBench (ours)} & \checkmark & SMR \& IMR & \checkmark & MEI & MLI & \checkmark \\
\bottomrule
\end{tabular}%
}

\label{tab:related}
\end{table*}

In practice, missingness is rarely symmetric across modalities. Some modalities are systematically more fragile or costly than others. Audio channels may fail more frequently than text transcripts. In medical or sensor-rich scenarios, high-cost modalities may only be available for a small fraction of samples \cite{sun2024redcore,zhao2025mce,zhang2024multimodal}. 
This leads to \emph{imbalanced missing rates} (IMR), where certain modalities are observed much more often than others. IMR changes the effective exposure of each modality during training. As a result, learning can become biased toward dominant modalities. Under IMR, a single modality may receive a disproportionate share of gradient updates. This effect can produce biased representations even when task accuracy remains high \cite{sun2024redcore,zhao2025mce,wu2022characterizing}.
Figure~\ref{fig:motiv-ratio} illustrates the difference between shared missing rates (SMR) and IMR at the data level. Under SMR, all modalities exhibit similar empirical missing rates. Under IMR, modality-specific exposure differs substantially, even when the average missing rate is matched. 
Figure~\ref{fig:motiv-grad} shows a complementary effect at the optimization level. Using RedCore \cite{sun2024redcore} as a representative IMR-aware method on IEMOCAP, the figure demonstrates that, under IMR conditions, the language modality consistently drives larger parameter updates than the visual and acoustic modalities. This observation indicates that IMR can induce persistent gradient dominance within the model. Importantly, this behavior is not revealed by task metrics alone.
Although several recent methods explicitly consider IMR  \cite{sun2024redcore, zhao2025mce, shi2024passion}, many missing-modality handling approaches still assume shared or nearly symmetric missing patterns across modalities \cite{lian2023gcnet,zhao2021MMIN,fu2024sdr,wang2023incomplete}. More broadly, existing evaluations rarely distinguish SMR from IMR in a systematic manner. They also lack standardized tools for modality- and optimization-level analysis across missingness regimes.

At the benchmark level, prior efforts such as MultiBench \cite{liang2021multibench} and multimodal affective benchmarks \cite{wu2024deep,lin2026missmac} provide unified evaluation platforms but primarily focus on balanced or fixed corruption scenarios. As a result, current benchmarks mainly answer how accurate a model is under missing modalities. They offer limited insight into which modality is disadvantaged or how strongly each modality influences the optimization process. No existing benchmark simultaneously distinguishes shared and imbalanced missing-rate regimes and provides modality-aware diagnostics beyond task performance. In realistic deployments, however, imbalanced missing rates are common, making such analysis necessary.

To address this gap, we introduce \textbf{MissBench}, a benchmark and framework for incomplete multimodal affective computing that explicitly accounts for both shared and imbalanced missing rates. MissBench reorganizes three English datasets (CMU-MOSI \cite{zadeh-mosi}, CMU-MOSEI \cite{zadeh2018mosei}, and IEMOCAP \cite{busso2008iemocap}) and one Chinese dataset (CH-SIMS \cite{yu2020ch}) under controlled masking protocols that cover SMR and a family of IMR settings. Beyond conventional task metrics, we propose two diagnostic measures. The \emph{Modality Equity Index} (MEI) quantifies how evenly different modalities contribute to predictive performance across missing-modality configurations. The \emph{Modality Learning Index} (MLI) captures optimization imbalance by summarizing modality-specific gradient dynamics during training.  MissBench exposes modality inequity and optimization dominance that remain hidden when evaluation is limited to task-level metrics.
Overall, our main contributions are:
\begin{itemize}
    \item We present \textbf{MissBench}, a benchmark for multimodal affective computing under incomplete modalities that standardizes shared and imbalanced missing-rate protocols on multiple datasets with fixed data splits and masking seeds for reproducible evaluation.
    \item We introduce two diagnostic metrics, the \textbf{Modality Equity Index} (MEI) and the \textbf{Modality Learning Index} (MLI), which quantify modality-level contribution equity and optimization imbalance from performance shifts and modality-specific gradient dynamics.
    \item We provide a unified evaluation pipeline and conduct a comprehensive empirical study across representative model families, showing that MissBench reveals modality inequity and optimization imbalance that are not captured by task metrics alone.
\end{itemize}

\section{Related Work}
\label{sec:related}

\textit{Multimodal affective computing and robustness.}
Multimodal affective computing has been extensively studied, with surveys summarizing advances in feature design, fusion strategies, and robustness to noise and distribution shifts \cite{zadeh-mosi,zadeh2018mosei,busso2008iemocap,hu2024recent}.
These works primarily aim to improve task performance under clean or mildly corrupted inputs.
When missing data are considered, they are typically treated as a secondary robustness factor rather than a central design axis.
As a result, robustness is often assessed only at the output level.
Modality-level contribution and optimization behavior remain largely unexplored in these studies.

\begin{figure*}
    \centering
    \includegraphics[width=\linewidth]{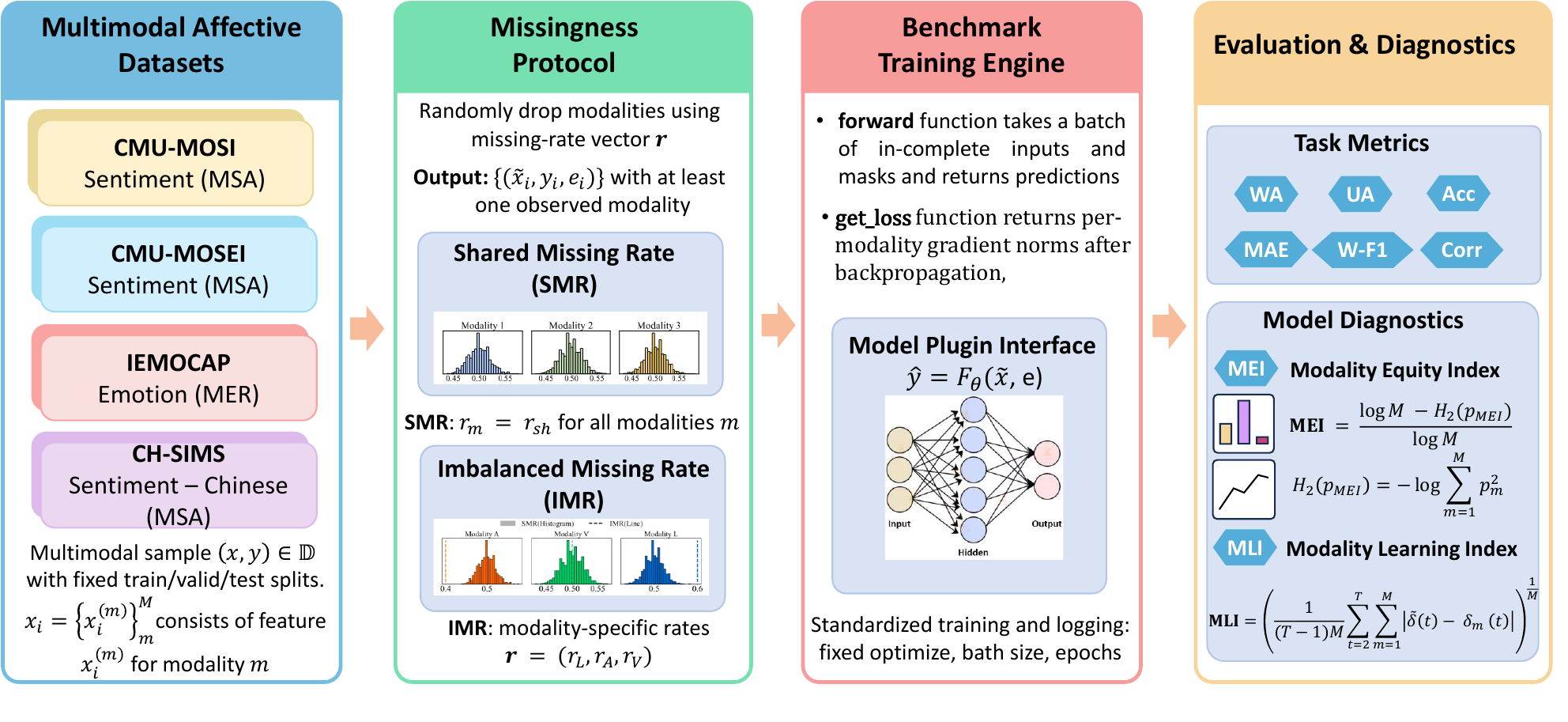}
    \caption{MissBench benchmark pipeline for multimodal affective tasks, illustrating data preparation, SMR/IMR missingness protocols, unified training with model plugins, and evaluation with both task metrics and modality-aware diagnostics.}
    \label{fig:missbench-pipeline}
\end{figure*}

\textit{Incomplete multimodal learning and missing modalities.}
A large body of methods has been proposed to address missing modalities.
Representative approaches include imagination-based reconstruction, graph-based completion, and generative or robust architectures for incomplete multimodal inputs \cite{ma2021smil,zhao2021MMIN,lian2023gcnet,wang2023incomplete,xu2024leveraging}.
These methods significantly improve robustness compared to naive fusion.
However, most of them assume fixed or shared missing patterns and are evaluated on a limited set of datasets.
Evaluation is predominantly conducted using task-level metrics such as accuracy, F1-score, and MAE.
In particular, the effect of systematically imbalanced missing rates on modality equity and learning dynamics is not explicitly analyzed.

\textit{Benchmarks and multimodal imbalance.}
Several benchmarks have been proposed to evaluate robustness in multimodal learning.
MultiBench \cite{liang2021multibench} and related efforts provide broad evaluations across tasks and domains \cite{wang2025umu,lian2026merbench,wu2024deep}.
These benchmarks mainly focus on balanced corruption or generic robustness scenarios.
They do not explicitly distinguish shared and imbalanced missing-rate regimes.
They also lack modality-aware diagnostic metrics beyond task performance.
Recent work on multimodal imbalance and IMR-aware learning, such as BalanceBenchmark \cite{xu2025balancebenchmark}, RedCore \cite{sun2024redcore}, and MCE \cite{zhao2025mce}, studies unequal modality contributions and imbalanced missing rates.
However, these efforts either target general multimodal tasks without affective-specific protocols or adopt method-specific setups that hinder direct comparison.
MissMAC-Bench~\cite{lin2026missmac} introduces a benchmark for missing modality issues in multimodal affective computing.
It evaluates holistic robustness under fixed and random missing patterns using competence-resilience measures.
However, it does not explicitly model imbalanced missing rates or provide modality-level or optimization-level diagnostics.

As summarized in Table~\ref{tab:related}, existing methods and benchmarks leave a clear gap: none simultaneously standardizes shared and imbalanced missing-rate protocols for multimodal affective tasks while exposing both modality-level contributions and optimization dynamics. As a result, evaluations may report similar task accuracy while masking substantial differences in modality dominance and learning imbalance. MissBench is designed to fill this gap.

\section{MissBench Design}
 Figure~\ref{fig:missbench-pipeline} summarizes the four phases of MissBench, including dataset organization, missingness protocol generation, standardized training, and modality-aware evaluation.

\subsection{Datasets and Tasks}
\label{sec:datasets_tasks}


\textbf{Datasets.}
Let $\mathbb{D} = \{(x_i, y_i)\}_{i=1}^{N}$ be a multimodal dataset with $M$ modalities and label space $\mathcal{Y}$.
Each input $x_i = \{x_i^{(m)}\}_{m=1}^{M}$ consists of features $x_i^{(m)} \in \mathbb{R}^{d_m}$ for modality $m$.
MissBench adopts four affective datasets with three core modalities: language (L), visual (V), and acoustic (A)
, namely IEMOCAP \cite{busso2008iemocap}, CMU-MOSI \cite{zadeh-mosi}, CMU-MOSEI \cite{zadeh2018mosei}, and CH-SIMS \cite{yu2020ch}.

After applying a masking protocol (Section~\ref{sec:missing_protocols}), each clean input $x_i$ is transformed into an incomplete sample $\tilde{x}_i = \{\tilde{x}_i^{(m)}\}_{m=1}^{M}$, forming the corrupted dataset $\tilde{\mathbb{D}} = \{(\tilde{x}_i, y_i)\}_{i=1}^{N}$.
We always ensure that at least one modality remains observed in every sample so that supervision is well defined.

\textbf{Task formulation.}
Given an incomplete sample $\tilde{x}_i$, a multimodal model $\mathrm{F}_\theta$ produces a prediction $\hat{y}_i = \mathrm{F}_\theta(\tilde{x}_i)$.
MissBench considers two task families.

\textit{Multimodal Emotion Recognition (MER).} On IEMOCAP, the label space is a set of emotion categories \(\mathcal{Y}_{\mathrm{MER}} = \{1,\dots,C\}\). The training objective is multiclass classification with cross-entropy loss:
\begin{equation}
\mathcal{L}_{\mathrm{MER}}(\theta) = \frac{1}{N} \sum_{i=1}^{N}
  \ell_{\mathrm{CE}}(\hat{y}_i, y_i)
\end{equation}
We report weighted accuracy (WA), unweighted accuracy (UA), and weighted F1-score (W-F1) as MER metrics.

\textit{Multimodal Sentiment Analysis (MSA).}
On CMU-MOSI, CMU-MOSEI, and CH-SIMS, each utterance is annotated with a real-valued sentiment intensity label.
Let $y_i^{\mathrm{reg}} \in \mathbb{R}$ denote the intensity label.
The model outputs $\hat{y}_i^{\mathrm{reg}}$, we use the following regression loss:
\begin{equation}
\mathcal{L}_{\mathrm{MSA}}(\theta)
= \frac{1}{N} \sum_{i=1}^{N}
    \ell_{\mathrm{reg}}(\hat{y}_i^{\mathrm{reg}}, y_i^{\mathrm{reg}})
\end{equation}
where $\ell_{\mathrm{reg}}$ is a regression loss (e.g., MAE or MSE).
For evaluation, we follow common practice and report binary accuracy (Acc-2), weighted F1-score (WAF), mean absolute error (MAE), and Pearson correlation (Corr).

Throughout this paper, training under SMR and IMR corresponds to minimizing the same task loss (MER or MSA) with masks drawn from $p_{\mathrm{SMR}}$ and $p_{\mathrm{IMR}}$, respectively.
Section~\ref{sec:metrics} introduces MEI and MLI to measure how these different masking distributions lead to inequity in modality contribution and optimization dynamics.

\subsection{Missingness Protocol}
\label{sec:missing_protocols}

Given a clean multimodal dataset $\mathbb{D}$, MissBench generates incomplete samples by applying a stochastic masking operator $\mathcal{M}(\cdot; \mathbf{r})$ parameterized by a missing-rate vector $\mathbf{r}$.

\textit{Masking operator.}
For each sample $x_i = \{x_i^{(m)}\}_{m=1}^{M}$ and missing-rate vector $\mathbf{r} = [r_1,\dots,r_M]$, the operator draws a binary mask $e_i = \{e_i^{(m)}\}_{m=1}^{M}$ with $e_i^{(m)} \in \{0,1\}$, $\sum_{m} e_i^{(m)} \ge 1$.
The incomplete sample $\tilde{x}_i^{(m)}$ is constructed as $\tilde{x}_i^{(m)} = x_i^{(m)}$ if $e_i^{(m)} = 1$, and $\vec{0}$ otherwise. MissBench instantiates two families of masking distributions, corresponding to shared and imbalanced missing-rate regimes.

\textit{Shared Missing Rate (SMR)}
Under SMR, all modalities share the same missing probability $r_{\mathrm{sh}} \in [0,1)$.
Each modality $m$ in sample $i$ is independently retained with probability $1 - r_{\mathrm{sh}}$ and dropped with probability $r_{\mathrm{sh}}$:
\begin{equation} 
\Pr(e_i^{(m)} = 1) = 1 - r_{\mathrm{sh}}, \quad
\Pr(e_i^{(m)} = 0) = r_{\mathrm{sh}}
\end{equation}
The induced distribution over a masking pattern
$e = \{e^{(m)}\}_{m=1}^{M}$ is:
\begin{equation}
p_{\mathrm{SMR}}(e)
= \frac{\prod_{m=1}^{M} (1 - r_{\mathrm{sh}})^{e^{(m)}}
                        r_{\mathrm{sh}}^{1 - e^{(m)}}}
       {1 - r_{\mathrm{sh}}^{M}}
\end{equation}
where the denominator excludes the all-missing case.
Under this regime, the empirical missing ratios
$\hat{r}_m = \frac{1}{N}\sum_{i=1}^{N}(1 - e_i^{(m)})$ concentrate around $r_{\mathrm{sh}}$ as $N$ grows, ensuring that all modalities are statistically symmetric and any observed differences arise from random
fluctuation rather than systematic imbalance.

\textit{Imbalanced Missing Rate (IMR)}
In realistic scenarios, different modalities have heterogeneous missing probabilities $r_m \neq r_{m'}$.
MissBench models this by assigning each modality its own missing ratio $r_m \in [0,1)$ and defining:
\begin{equation}
p_{\mathrm{IMR}}(e)
= \frac{\prod_{m=1}^{M} (1 - r_m)^{e^{(m)}} r_m^{1 - e^{(m)}}}
       {1 - \prod_{m=1}^{M} r_m}
\end{equation}
Compared with a mean-matched SMR distribution, $p_{\mathrm{IMR}}$ concentrates probability mass on patterns that retain low-$r_m$ modalities while dropping high-$r_m$ ones, creating systematic exposure imbalance across modalities.
The degree of imbalance can be quantified by a divergence:
$
\Delta_{\mathrm{IMR}}
= \mathcal{D}\big(p_{\mathrm{IMR}}(e) \,\|\, p_{\mathrm{SMR}}(e)\big)
$,
where $\mathcal{D}$ can be, for example, KL or Jensen–Shannon divergence. Larger $\Delta_{\mathrm{IMR}}$ indicates stronger modality heterogeneity and
a more biased sampling of multimodal patterns.

\subsection{Metrics for Modality Equity and Optimization}
\label{sec:metrics}

Given a multimodal model $\mathrm{F}_\theta$ trained under a masking protocol (SMR or IMR), MissBench evaluates not only task performance but also the equity of modality contributions and the balance of optimization
dynamics.
We formalize $\mathrm{F}_\theta$ as consisting of $\mathcal{M}$ modules $\{f_{\theta_1},\dots,f_{\theta_\mathcal{M}}\}$, where each module $f_{\theta_i}$ typically corresponds to a modality encoder or a shared fusion component.
We introduce two diagnostic metrics: the \textit{Modality Equity Index (MEI)} and the \textit{Modality Learning Index (MLI)}.

\subsubsection{Modality Equity Index (MEI)}
It measures how evenly different modalities contribute to predictive performance by evaluating model behavior across all possible missing-modality configurations.

To compute MEI, we evaluate the trained model on the uncorrupted test dataset $\mathcal{S}$ under synthetic missing-modality patterns.
For each modality $m$, let $\mathcal{C}_m$ denote the set of all modality combinations in which $m$ is unavailable, with $|\mathcal{C}_m|=2^{M-1}-1$.
For a specific combination $\hat{c} \in \mathcal{C}_m$, we construct a synthetic test set $\mathcal{S}_{\hat{c}}$ by replacing features of modalities not in $\hat{c}$ with zero vectors $\vec{0}$.
Let $\mathrm{Perf}_{\hat{c}}$ be the task-specific performance score of $\mathrm{F}_\theta$ on $\mathcal{S}_{\hat{c}}$.

We first measure the performance fluctuation associated with modality $m$ across all combinations in $\mathcal{C}_m$:
\begin{equation}
    \mathbf{s}_m = \{\mathrm{Perf}_{\text{full}}-\mathrm{Perf}_{\hat{c}} : \hat{c}\in\mathcal{C}_m\}
\end{equation}
where $\mathbf{s}_m\in\mathbb{R}^{|\mathcal{C}_m|}$ captures the performance drop when $m$ is removed in each combination.
The contribution of modality $m$ is characterized by the mean $\mu_m = \mathbb{E}[\mathbf{s}_m]$ and standard deviation $\sigma_m = \sqrt{\text{Var}(\mathbf{s}_m)}$.

We then normalize the contribution as:
\begin{equation}
    \varsigma_m =\frac{\mu_m}{\sigma_m+\varepsilon}
\end{equation}
where $\varepsilon > 0$ ensures numerical stability.
This normalization yields a signal-to-noise ratio: modalities with consistent, large performance drops have high $|\varsigma_m|$, while those with noisy or small drops have low $|\varsigma_m|$.

To construct a probability distribution over modality contributions, we define:
\begin{equation}
    p_m = \frac{|\varsigma_m|}{\sum_{m'=1}^M |\varsigma_{m'}| + \varepsilon}, \quad m = 1,\ldots,M
\end{equation}
and let $\mathbf{p}_{\mathrm{MEI}} = [p_1, p_2, \ldots, p_M]$ represent the normalized contribution distribution.
Here, we use the magnitude $|\varsigma_m|$ to capture the strength of modality $m$'s contribution regardless of sign, as large absolute values indicate strong impact (beneficial if $\varsigma_m > 0$, detrimental if $\varsigma_m < 0$).

Finally, MEI is defined as:
\begin{equation}
    \mathrm{MEI}=\frac{\log M - H_2(\mathbf{p}_{\mathrm{MEI}})}{\log M}
\end{equation}
where $H_2(\mathbf{p}) = -\log \sum_{m=1}^M p_m^2$ is the Rényi entropy of order 2.
MEI ranges from 0 (one modality dominates) to 1 (perfectly balanced contribution).
High MEI indicates that all modalities contribute evenly, while low MEI signals that one or a few modalities dominate task performance.

\subsubsection{Modality Learning Index (MLI)}
MLI quantifies optimization imbalance by comparing gradient magnitudes
across modalities during training.

Let $\mathcal{L}^{(t)}=\frac{1}{N}\sum_{i=1}^{N}\ell^{(t)}_i$ be the task loss at iteration $t$.
We define the \emph{modality-specific loss} for modality $m$ as the average loss over samples in which $m$ is available:
\begin{equation}
    \mathcal{L}^{(t)}_m = \frac{\sum_{i=1}^{N}e^{(m)}_i\ell^{(t)}_i}{\sum_{i=1}^{N}e^{(m)}_i}
\end{equation}
where $e^{(m)}_i \in \{0,1\}$ indicates whether modality $m$ is observed in sample $i$.
For each module $f_{\theta_k}$, we compute its modality-specific gradient vector at step $t$ via:
\begin{equation}
    \mathbf{g}^m_{\theta_k}(t) = \frac{\partial\mathcal{L}_m^{(t)}}{\partial\theta_k}
\end{equation}
The optimization magnitude for modality $m$ is then estimated by averaging gradient norms across all $\mathcal{M}$ modules:
\begin{equation}
    \mathrm{G}_m(t) = \frac{1}{\mathcal{M}}\sum_{k=1}^{\mathcal{M}}\|\mathbf{g}^m_{\theta_k}(t)\|_2
\end{equation}
This aggregation provides a holistic view of how strongly modality $m$ drives parameter updates across the entire model.

We define the temporal gradient variation of modality $m$ at training step $t$ as:
\begin{equation}
    \delta_m(t) = |\mathrm{G}_m(t) - \mathrm{G}_m(t-1)|,\quad t=2,\dots,T
\end{equation}
which quantifies the change in the magnitude of the uni-modal gradient between consecutive iterations. $T$ denotes the total number of training iterations, and $\delta_m(t)$ is defined from the second iteration onward. To capture the overall update behavior across modalities, we further compute the cross-modal variation:
\begin{equation}
    \bar{\delta}(t) = \frac{1}{M}\sum_{m=1}^M \delta_m(t)
\end{equation}

Finally, our metric MLI is formulated as:
\begin{equation}
    \mathrm{MLI} = \Bigg(\frac{1}{\max_t\bar{\delta}(t)(T-1)M}\sum_{t=2}^T\sum_{m=1}^M|\bar{\delta}(t) - \delta_m(t)|\Bigg)^{\frac{1}{M}}
\end{equation}
Overall, MLI measures the extent to which individual modalities deviate from the shared update dynamics throughout training. $\mathrm{MLI}\in[0,1]$, where lower values indicate more consistent and balanced temporal updates across modalities, whilst higher values reflect heterogeneous update stability, suggesting asynchronous optimization during learning.

\subsection{Benchmark Pipeline and Model Plugin}
\label{sec:pipeline}

MissBench provides a unified pipeline that applies missingness protocols, trains models, and evaluates arbitrary methods through a simple plugin interface, ensuring fair comparison under identical masking, training, and evaluation conditions.

\textit{Overall pipeline.}
Given a clean dataset $\mathbb{D}$, MissBench first applies a chosen missingness protocol (SMR or IMR) to generate the corrupted dataset $\tilde{\mathbb{D}} = \{(\tilde{x}_i, y_i)\}_{i=1}^{N}$ together with binary masks $e_i$ that indicate which modalities remain observed in each sample.
During training, the benchmark repeatedly samples mini-batches $\{(\tilde{x}_i, e_i, y_i)\}$, feeds them to a user-specified model plugin $\mathrm{F}_\theta$, and optimizes the 
total objective. The pipeline supports flexible optimization by aggregating the task loss (MER or MSA) with any model-specific auxiliary losses returned by the plugin. At evaluation time, task metrics are computed on held-out data, and MEI and MLI are derived from logged performance and gradient statistics.

\textit{Model plugin interface.}
To integrate a new method, users implement a model plugin that exposes two functions: \texttt{forward} and \texttt{get\_loss}.
The \texttt{forward} function takes a batch of incomplete inputs and masks as input, and returns predictions and intermediate representations:
\begin{equation}
\hat{y}, {z} = \mathrm{F}_\theta(\tilde{x}, e)
\end{equation}
where ${z}$ denotes optional auxiliary outputs and $e$ allows the model to identify missing modalities without accessing ground-truth full features.
The \texttt{get\_loss} function then consumes these outputs to compute auxiliary objectives, allowing the benchmark to support complex multi-task learning frameworks. By offloading diagnostics to external \textbf{MLI} and \textbf{MEI} modules, we eliminate the need for complex internal hooks within the model.

\textit{Standardized training and logging.}
All methods are trained under the same optimization budget: identical batch size, optimizer, number of epochs, and early-stopping criterion. This ensures that differences in results arise from modeling choices rather than hyperparameter tuning advantages. During training, MissBench logs per-epoch task metrics, modality-wise performance on validation data, and uses the MLI evaluator to track the fluctuation of gradient norms for each modality. These logs enable the computation of MLI over the training trajectory. Complementing this, MEI is evaluated through a separate ablation inference loop on the valid and test set, and these metrics jointly provide a consistent diagnostic view of how each model behaves under SMR and IMR protocols.

\begin{table*}[t!]
    \centering
    \caption{$\text{MEI}_{\text{UA}}$/MLI(\%) of representatives from each method-family.} 
    \label{tab:mei-mli-merge}
    \resizebox{\linewidth}{!}{%
    \begin{tabular}{llcccccc|cccccc}
\toprule
\multirow{3}{*}{Method} &
\multirow{3}{*}{Setting} &

\multicolumn{6}{c}{IEMOCAP} &
\multicolumn{6}{c}{CMU-MOSI} \\

\cmidrule(lr){3-8} \cmidrule(lr){9-14}

& &
\multicolumn{2}{c}{0.3 / (0.1,0.2,0.6)} &
\multicolumn{2}{c}{0.5 / (0.2,0.5,0.8)} &
\multicolumn{2}{c}{0.7 / (0.4,0.8,0.9)} &

\multicolumn{2}{c}{0.3 / (0.1,0.2,0.6)} &
\multicolumn{2}{c}{0.5 / (0.2,0.5,0.8)} &
\multicolumn{2}{c}{0.7 / (0.4,0.8,0.9)} \\

\cmidrule(lr){3-4}\cmidrule(lr){5-6}\cmidrule(lr){7-8}
\cmidrule(lr){9-10}\cmidrule(lr){11-12}\cmidrule(lr){13-14}

& &
MEI$_{UA}$ & MLI &
MEI$_{UA}$ & MLI &
MEI$_{UA}$ & MLI &

MEI$_{UA}$ & MLI &
MEI$_{UA}$ & MLI &
MEI$_{UA}$ & MLI \\

\midrule

\multirow{2}{*}{GCNet}
& SMR
& $8.21_{\pm5.40}$ & $31.56_{\pm3.08}$
& $11.41_{\pm8.31}$ & $38.83_{\pm6.99}$
& $20.57_{\pm17.78}$ & $41.15_{\pm6.45}$

& $79.67_{\pm14.41}$ & $34.77_{\pm7.29}$
& $76.67_{\pm7.15}$ & $43.12_{\pm8.51}$
& $48.75_{\pm30.61}$ & $62.42_{\pm5.19}$ \\

& IMR
& $7.33_{\pm2.48}$ & $32.89_{\pm3.55}$
& $30.71_{\pm34.82}$ & $42.52_{\pm8.69}$
& $25.06_{\pm28.70}$ & $41.97_{\pm4.62}$

& $70.83_{\pm10.94}$ & $41.06_{\pm9.81}$
& $71.24_{\pm18.21}$ & $55.85_{\pm10.38}$
& $79.90_{\pm15.55}$ & $58.02_{\pm9.19}$ \\

\midrule

\multirow{2}{*}{RedCore}
& SMR
& $5.92_{\pm2.71}$ & $34.77_{\pm7.29}$
& $13.45_{\pm8.89}$ & $44.03_{\pm2.64}$
& $11.48_{\pm10.00}$ & $43.12_{\pm8.51}$

& $88.08_{\pm2.50}$ & $34.28_{\pm2.56}$
& $85.04_{\pm3.57}$ & $41.76_{\pm5.30}$
& $83.57_{\pm8.65}$ & $49.20_{\pm5.40}$ \\

& IMR
& $7.06_{\pm1.73}$ & $41.06_{\pm9.81}$
& $8.39_{\pm10.67}$ & $46.81_{\pm4.34}$
& $5.63_{\pm2.50}$ & $55.85_{\pm10.38}$

& $86.85_{\pm2.03}$ & $40.39_{\pm5.54}$
& $87.16_{\pm2.13}$ & $40.86_{\pm7.70}$
& $80.58_{\pm6.82}$ & $51.01_{\pm2.50}$ \\

\midrule

\multirow{2}{*}{Ada2I}
& SMR
& $10.65_{\pm6.85}$ & $32.87_{\pm1.71}$
& $6.77_{\pm7.77}$ & $38.00_{\pm3.35}$
& $10.21_{\pm8.54}$ & $39.50_{\pm5.47}$

& $67.60_{\pm18.78}$ & $32.87_{\pm2.79}$
& $60.40_{\pm17.71}$ & $38.85_{\pm3.91}$
& $42.91_{\pm22.21}$ & $39.50_{\pm5.13}$ \\

& IMR
& $10.58_{\pm12.20}$ & $35.48_{\pm3.27}$
& $9.68_{\pm8.71}$ & $41.73_{\pm6.45}$
& $20.86_{\pm25.99}$ & $43.69_{\pm2.07}$

& $40.47_{\pm19.05}$ & $35.48_{\pm4.47}$
& $50.97_{\pm36.04}$ & $38.16_{\pm4.37}$
& $49.23_{\pm14.34}$ & $43.69_{\pm2.85}$ \\

\bottomrule
\end{tabular}
    }
    
\end{table*}
\section{MissBench Empirical Analysis}
\label{sec:analysis}
\subsection{Experimental Setup}



We evaluate MissBench on four multimodal affective datasets (IEMOCAP \cite{busso2008iemocap}, CMU-MOSI \cite{zadeh-mosi}, CMU-MOSEI \cite{zadeh2018mosei}, CH-SIMS \cite{yu2020ch}) with language, acoustic, and visual modalities, following the standard MER/MSA protocols and official splits when available.
Compared methods cover three families: (i) IMR-aware approaches (RedCore \cite{sun2024redcore}, MCE \cite{zhao2025mce}), (ii) missing-modality handling models (MMIN \cite{zhao2021MMIN}, GCNet \cite{lian2023gcnet}, SDR-GNN \cite{fu2024sdr}, Mi-CGA \cite{nguyen2025-micga}), and (iii) gradient-based or generic baselines (Ada2I \cite{nguyen2024ada2i}, OGM-GE \cite{peng2022balanced}, naive fusion).

We evaluate all methods under identical training budgets (optimizer, batch size, number of epochs) and report results for both SMR and IMR masking protocols with fixed random seeds to ensure fair and reproducible comparisons.
Our empirical study is organized around the following research questions:

\textbf{RQ1} (\emph{Shared missing rates}). Do existing methods behave as expected under standard shared missing rates (SMR), in terms of both task performance and modality/gradient diagnostics?

\textbf{RQ2} (\emph{Mean-matched IMR}). When the average missing rate remains constant, how does moving from SMR to imbalanced missing rates (IMR) affect performance, MEI, and MLI?

\textbf{RQ3} (\emph{Extreme IMR and trade-offs}). Under more extreme IMR scenarios, how do different method families trade off task accuracy against modality equity (MEI) and optimization balance (MLI)?

\subsection{Behavior under Shared Missing Rates (SMR)}

\begin{figure}[t!]
    \centering
    \includegraphics[width=\linewidth]{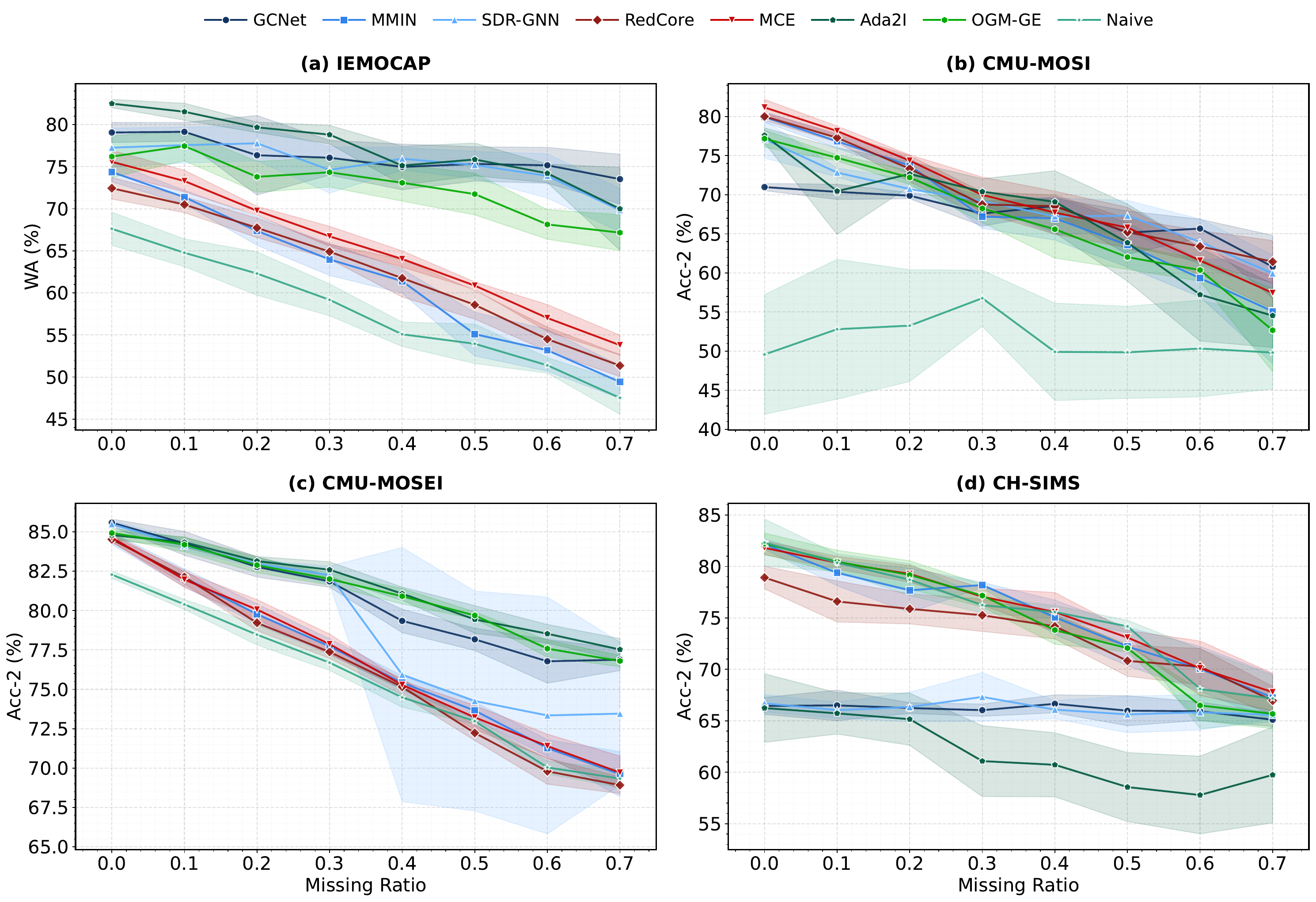}
    \caption{Performance of multimodal methods under shared missing rates (SMR) on IEMOCAP (WA) and CMU-MOSI/MOSEI/CH-SIMS (Acc-2). IMR-aware methods are shown in red, missing-modality handling models in blue, and gradient-based or generic baselines in green.}
    \label{fig:smr-curves}
\end{figure}

\textit{Task-level performance under SMR.}  We address \textbf{RQ1} by examining model behavior under shared missing rates (SMR). Figure~\ref{fig:smr-curves} shows that, across all datasets, task performance degrades smoothly and monotonically as the shared missing ratio increases, confirming that SMR induces a controlled and predictable degradation. IMR-aware models and dedicated missing-modality handlers consistently outperform naive fusion and generic baselines, with more pronounced gains at moderate to high missing ratios $(0.3\text{–}0.7)$. 

However, similar task-level trends may arise from fundamentally different modality usage patterns, which cannot be distinguished by accuracy alone.

\textit{MEI and MLI under SMR.}
Table~\ref{tab:mei-mli-merge} reveals substantial dataset-dependent differences in modality equity and optimization balance under SMR. On IEMOCAP, all models exhibit low $\mathrm{MEI}_{\mathrm{UA}}$ (mostly below $25$) together with relatively high MLI values, indicating pronounced modality and optimization imbalance even under symmetric missing patterns. In contrast, on CMU-MOSI, the same models achieve much higher $\mathrm{MEI}_{\mathrm{UA}}$ (typically $70$--$90$) with more moderate MLI, suggesting more balanced modality exploitation.

Overall, while SMR provides a stable baseline with smooth task degradation, modality equity and learning balance vary substantially across datasets. Notably, IEMOCAP already exhibits strong imbalance under SMR, motivating the more challenging IMR settings examined next.
\begin{table*}[t!] 
\centering 
\caption{Task performance on CMU-MOSI under SMR and mean-matched IMR across low, medium, and high missing regimes. For IMR, modality-specific rates follow the order $(r_A, r_V, r_L)$.} 
\label{tab:smr-imr-mosi} 
\resizebox{0.85\linewidth}{!}{%
\begin{tabular}{llccccccccc} \toprule \multirow{2}{*}{Method} & \multirow{2}{*}{Setting} & \multicolumn{3}{c}{0.3 / (0.1, 0.2, 0.6)} & \multicolumn{3}{c}{0.5 / (0.2, 0.5, 0.8)} & \multicolumn{3}{c}{0.7 / (0.4, 0.8, 0.9)} \\ \cmidrule(lr){3-5} \cmidrule(lr){6-8} \cmidrule(lr){9-11} & & Acc-2($\uparrow$) & MAE($\downarrow$) & Corr($\uparrow$) & Acc-2($\uparrow$) & MAE($\downarrow$) & Corr($\uparrow$) & Acc-2($\uparrow$) & MAE($\downarrow$) & Corr($\uparrow$) \\ \midrule \multirow{2}{*}{MMIN} & SMR & $67.20_{\pm 1.48}$ & $1.180_{\pm 0.085}$ & $0.543_{\pm 0.025}$ & $63.57_{\pm 2.82}$ & $1.249_{\pm 0.008}$ & $0.472_{\pm 0.015}$ & $55.09_{\pm 6.66}$ & $1.374_{\pm 0.064}$ & $0.331_{\pm 0.092}$ \\ & IMR & $61.80_{\pm 1.43}$ & $1.288_{\pm 0.017}$ & $0.425_{\pm 0.020}$ & $55.61_{\pm 3.86}$ & $1.398_{\pm 0.042}$ & $0.305_{\pm 0.064}$ & $52.59_{\pm 4.11}$ & $1.425_{\pm 0.027}$ & $0.293_{\pm 0.027}$ \\ \midrule \multirow{2}{*}{GCNet} & SMR & $67.59_{\pm 1.60}$ & $1.227_{\pm 0.025}$ & $0.453_{\pm 0.014}$ & $65.15_{\pm 2.77}$ & $1.283_{\pm 0.057}$ & $0.428_{\pm 0.020}$ & $60.79_{\pm 4.01}$ & $1.387_{\pm 0.043}$ & $0.284_{\pm 0.070}$ \\ & IMR & $64.33_{\pm 3.04}$ & $1.316_{\pm 0.055}$ & $0.346_{\pm 0.073}$ & $55.58_{\pm 3.78}$ & $1.436_{\pm 0.043}$ & $0.167_{\pm 0.066}$ & $55.95_{\pm 4.01}$ & $1.425_{\pm 0.058}$ & $0.208_{\pm 0.043}$ \\ \midrule \multirow{2}{*}{RedCore} & SMR & $68.75_{\pm 1.59}$ & $1.188_{\pm 0.032}$ & $0.509_{\pm 0.013}$ & $65.21_{\pm 1.73}$ & $1.256_{\pm 0.008}$ & $0.433_{\pm 0.023}$ & $61.43_{\pm 2.70}$ & $1.347_{\pm 0.047}$ & $0.352_{\pm 0.047}$ \\ & IMR & $62.87_{\pm 2.14}$ & $1.307_{\pm 0.029}$ & $0.393_{\pm 0.033}$ & $59.48_{\pm 2.09}$ & $1.372_{\pm 0.022}$ & $0.323_{\pm 0.029}$ & $54.60_{\pm 5.43}$ & $1.424_{\pm 0.038}$ & $0.258_{\pm 0.117}$ \\ \midrule \multirow{2}{*}{MCE} & SMR & $69.97_{\pm 2.29}$ & $1.146_{\pm 0.041}$ & $0.547_{\pm 0.015}$ & $65.79_{\pm 2.60}$ & $1.230_{\pm 0.028}$ & $0.467_{\pm 0.015}$ & $57.44_{\pm 2.06}$ & $1.357_{\pm 0.037}$ & $0.367_{\pm 0.037}$ \\ & IMR & $63.35_{\pm 1.49}$ & $1.300_{\pm 0.029}$ & $0.413_{\pm 0.021}$ & $58.05_{\pm 1.23}$ & $1.381_{\pm 0.017}$ & $0.323_{\pm 0.034}$ & $57.96_{\pm 1.49}$ & $1.367_{\pm 0.015}$ & $0.335_{\pm 0.014}$ \\ \midrule \multirow{2}{*}{Ada2I} & SMR & $70.40_{\pm 1.65}$ & $1.380_{\pm 0.097}$ & $0.494_{\pm 0.027}$ & $63.87_{\pm 4.92}$ & $1.387_{\pm 0.097}$ & $0.362_{\pm 0.091}$ & $54.54_{\pm 4.07}$ & $1.472_{\pm 0.065}$ & $0.158_{\pm 0.089}$ \\ & IMR & $57.47_{\pm 3.24}$ & $1.494_{\pm 0.106}$ & $0.234_{\pm 0.093}$ & $55.12_{\pm 2.87}$ & $1.452_{\pm 0.042}$ & $0.174_{\pm 0.071}$ & $52.10_{\pm 3.94}$ & $1.552_{\pm 0.080}$ & $0.134_{\pm 0.072}$ \\ \midrule \multirow{2}{*}{OGM-GE} & SMR & $68.24_{\pm 1.80}$ & $1.234_{\pm 0.026}$ & $0.450_{\pm 0.030}$ & $62.02_{\pm 1.56}$ & $1.315_{\pm 0.029}$ & $0.360_{\pm 0.043}$ & $52.67_{\pm 5.31}$ & $1.444_{\pm 0.042}$ & $0.144_{\pm 0.099}$ \\ & IMR & $59.76_{\pm 3.13}$ & $1.375_{\pm 0.050}$ & $0.278_{\pm 0.073}$ & $53.38_{\pm 2.73}$ & $1.453_{\pm 0.031}$ & $0.138_{\pm 0.077}$ & $51.98_{\pm 3.65}$ & $1.468_{\pm 0.036}$ & $0.119_{\pm 0.052}$ \\ \bottomrule \end{tabular} %
} 
\end{table*}

\begin{table}[t!] \centering \caption{Task performance on IEMOCAP under SMR and mean-matched IMR across low, medium, and high missing regimes. IMR configurations are reported using the modality order $(r_A, r_V, r_L)$.} \label{tab:smr-imr-iemocap} \resizebox{\linewidth}{!}{%
\begin{tabular}{llccccccccc}
\toprule
\multirow{2}{*}{Method} &
\multirow{2}{*}{Setting} &
\multicolumn{2}{c}{0.3 / (0.1, 0.2, 0.6)} &
\multicolumn{2}{c}{0.5 / (0.2, 0.5, 0.8)} &
\multicolumn{2}{c}{0.7 / (0.4, 0.8, 0.9)} \\ 
\cmidrule(lr){3-4} \cmidrule(lr){5-6} \cmidrule(lr){7-8}
                         &     & WA    & F1    & WA    & F1    & WA    & F1 \\ 
\midrule
\multirow{2}{*}{MMIN}    & SMR & $63.98_{\pm 1.90}$ &    $64.30_{\pm 1.88}$   & $55.10_{\pm 2.60}$ &    $55.36_{\pm 2.67}$   & $49.43_{\pm 1.96}$ &   $49.51_{\pm 1.94}$  \\
                         & IMR & $61.00_{\pm 2.53}$ & $61.19_{\pm 2.57}$ & $55.55_{\pm 1.95}$ & $55.91_{\pm 1.91}$ & $50.25_{\pm 1.72}$ & $50.18_{\pm 2.03}$     \\
\midrule
\multirow{2}{*}{GCNet}   & SMR & $76.07_{\pm 2.06}$ &    $76.23_{\pm 1.95}$   & $75.33_{\pm 2.04}$ &    $75.49_{\pm 1.99}$   & $73.52_{\pm 2.95}$ &  $73.54_{\pm 2.89}$   \\
                         & IMR & $73.70_{\pm 3.58}$ &   $73.96_{\pm 3.26}$    & $65.21_{\pm 4.47}$ &   $65.35_{\pm 4.67}$    & $65.61_{\pm 3.70}$ &   $66.09_{\pm 3.69}$  \\
\midrule
\multirow{2}{*}{RedCore} & SMR & $64.90_{\pm 0.88}$ &   $65.15_{\pm 0.92}$    & $58.58_{\pm 1.69}$ &    $58.82_{\pm 1.66}$   & $51.36_{\pm 1.30}$ &   $51.71_{\pm 1.28}$  \\
                         & IMR & $62.69_{\pm 2.17}$ &    $62.72_{\pm 2.18}$   & $57.16_{\pm 1.58}$ &  $57.34_{\pm 1.60}$     & $52.38_{\pm 1.09}$ &   $52.53_{\pm 1.01}$  \\
\midrule
\multirow{2}{*}{MCE}     & SMR & $66.72_{\pm 1.20}$ &   $66.93_{\pm 1.18}$    & $60.87_{\pm 0.44}$ &   $61.06_{\pm 0.47}$    & $53.80_{\pm 1.19}$ &   $53.97_{\pm 1.16}$  \\
                         & IMR & $63.87_{\pm 1.16}$ &   $63.98_{\pm 1.16}$    & $59.27_{\pm 1.03}$ &   $59.48_{\pm 0.96}$    & $54.44_{\pm 1.38}$ &  $54.47_{\pm 1.37}$  \\
\midrule
\multirow{2}{*}{Ada2I}   & SMR & $78.81_{\pm 1.10}$ &    $78.81_{\pm 1.22}$   & $75.86_{\pm 1.95}$ &   $76.00_{\pm 1.89}$    & $69.99_{\pm 4.92}$ &   $70.26_{\pm 5.11}$  \\
                         & IMR & $76.52_{\pm 2.39}$ &    $76.67_{\pm 2.31}$   & $71.97_{\pm 2.14}$ &   $72.34_{\pm 2.18}$    & $71.94_{\pm 2.22}$ &   $72.40_{\pm 2.21}$  \\
\midrule
\multirow{2}{*}{OGM-GE}  & SMR & $74.34_{\pm 1.76}$  &    $74.62_{\pm 1.67}$   & $71.74_{\pm 2.47}$ &   $71.85_{\pm 2.25}$    & $67.16_{\pm 2.14}$ &   $67.24_{\pm 2.11}$ \\
                         & IMR & $70.93_{\pm 3.15}$ &    $71.11_{\pm 3.21}$   & $63.35_{\pm 3.89}$ &   $63.51_{\pm 3.80}$    & $63.05_{\pm 1.62}$ &    $63.72_{\pm 1.52}$ \\ 
\bottomrule
\end{tabular} %
} \end{table}

\subsection{Mean-matched SMR vs IMR}

\textit{Mean-matched protocol.}
To isolate the effect of imbalance from overall missing severity, we construct SMR--IMR pairs with matched expected missing ratios. Under SMR with rate $r_{\mathrm{sh}}$, the expected number of missing modalities per sample is $M r_{\mathrm{sh}}$, whereas under IMR with modality-specific rates $\mathbf{r} = (r_1,\dots,r_M)$ it is $\sum_{m=1}^{M} r_m$. We define the mean-matched condition as:
\begin{equation}
M r_{\mathrm{sh}} \approx \sum_{m=1}^{M} r_m .
\end{equation}
Under this constraint, SMR and IMR exhibit comparable overall missingness, and observed differences can be attributed to imbalance in modality exposure rather than total missing data.

We instantiate three mean-matched pairs corresponding to \textit{low}, \textit{medium}, and \textit{high} missing regimes with $r_{\mathrm{sh}} \in \{0.3, 0.5, 0.7\}$. Each SMR setting is paired with an IMR configuration whose $(r_A, r_V, r_L)$ values approximate the same average missing ratio. We assign higher missing rates to language to reflect its dominant role in many multimodal affective models; however, this choice is illustrative, and MissBench supports arbitrary SMR/IMR configurations and modality orderings.

\begin{figure}[t]
    \centering

    \begin{subfigure}{\linewidth}
        \centering
        \includegraphics[width=0.9\linewidth]{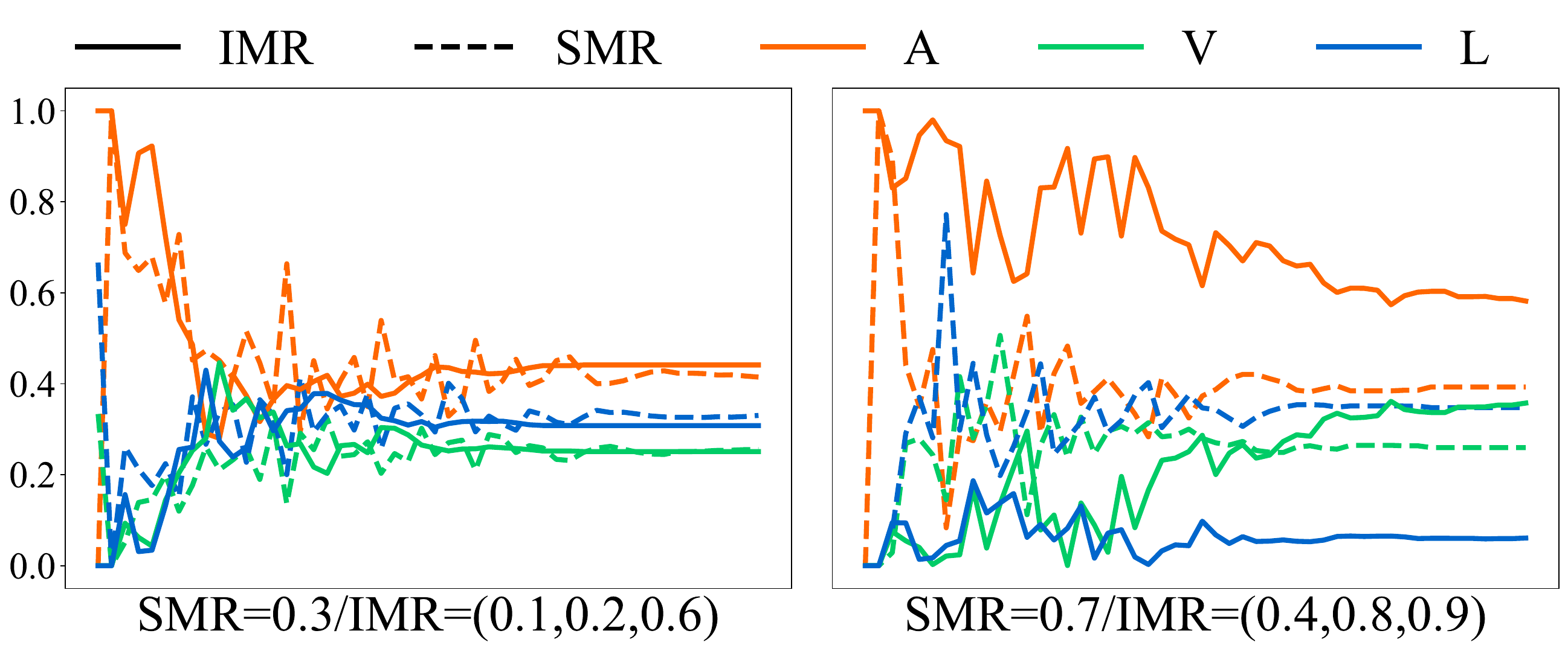}
        \subcaption{Modality contribution $\mathbf{p}_\mathrm{MEI}$ on validation data (by epoch)}

    \end{subfigure}


    \begin{subfigure}{\linewidth}
        \centering
        \includegraphics[width=0.9\linewidth]{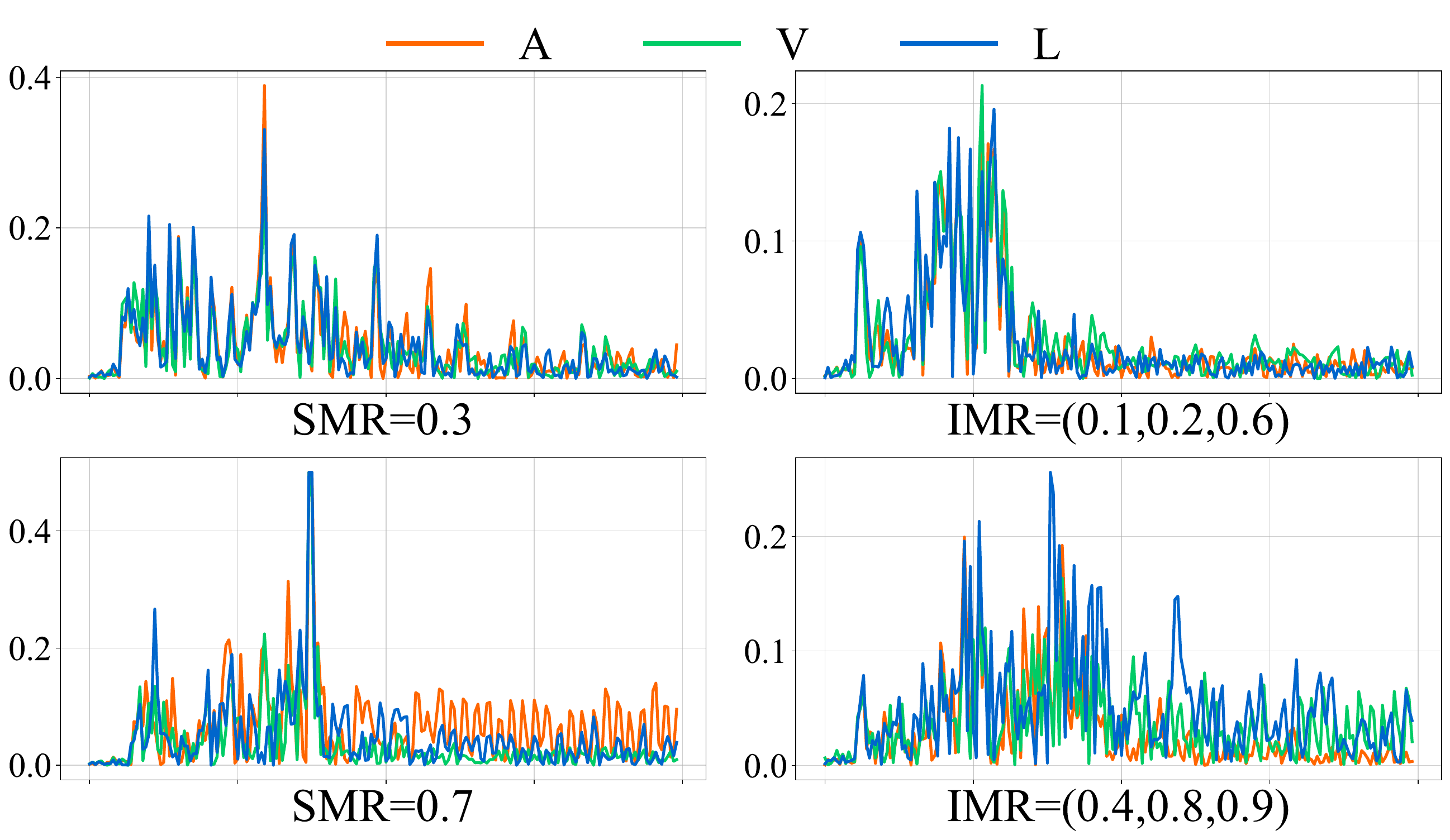}
        \subcaption{Unimodal temporal gradient variation $\delta_m(t)$ from MLI (by step)} 
    \end{subfigure}

    \caption{Statistics of GCNet \cite{lian2023gcnet} during training on IEMOCAP under different missing configurations.}
    \label{fig:GCNet-prob-grad}
\end{figure}

\textit{Task performance under mean-matched SMR vs.\ IMR. }
On CMU-MOSI (Table~\ref{tab:smr-imr-mosi}), moving from SMR to mean-matched IMR consistently degrades task performance across all method families. For missing-modality handlers such as MMIN and GCNet, Acc-2 drops by approximately $3\text{–}10$ points at the same mean missing level, accompanied by higher MAE and lower correlation across low, medium, and high regimes. IMR-aware methods also incur non-trivial losses. For example, RedCore’s Acc-2 decreases from $68.8$ to $62.9$ in the low regime and from $61.4$ to $54.6$ in the high regime, with corresponding increases in MAE and decreases in Corr. Generic or gradient-based baselines are the most sensitive. Ada2I’s Acc-2 drops from $70.4$ to $57.5$ under the low setting, and its correlation nearly halves from $0.494$ to $0.234$, with further degradation at higher missing levels.

On IEMOCAP (Table~\ref{tab:smr-imr-iemocap}), a similar pattern is observed, though with greater variation across methods. GCNet and OGM-GE show substantial WA drops when switching from SMR to IMR, particularly at medium and high missing levels. For instance, GCNet decreases from $75.3$ to $65.2$ at the medium setting and from $73.5$ to $65.6$ at the high setting. IMR-aware methods such as RedCore and MCE exhibit smaller but consistent degradations at low and medium levels, with partial stabilization at the highest missing ratio. Ada2I experiences relatively mild WA degradation and even a slight improvement at the highest setting (from $70.0$ to $71.9$), but its absolute performance remains below that of leading missing-modality handlers, and its instability on CMU-MOSI indicates limited robustness across tasks.

\begin{table*}[t!]
    \centering
    \caption{Performance comparison under different bimodal combinations (A: Audio, V: Visual, L: Language) and missing-rate settings, including Shared Missing Rate (SMR) and Imbalanced Missing Rate (IMR).
}
\label{tab:bimodal-result}
    \resizebox{\linewidth}{!}{\begin{tabular}{lccccccccccccc}
\toprule

\multirow{2}{*}{Method} 
& \multirow{2}{*}{\textbf{Setting (AV)}} 
& \multicolumn{4}{c}{0.1 / (0.1, 0.1)} 
& \multicolumn{4}{c}{0.3 / (0.1, 0.5)} 
& \multicolumn{4}{c}{0.5 / (0.6, 0.4)} \\

\cmidrule(lr){3-6}
\cmidrule(lr){7-10}
\cmidrule(lr){11-14}

& 
& WA & F1 & $\mathrm{MEI}_{\mathrm{UA}}$ & MLI
& WA & F1 & $\mathrm{MEI}_{\mathrm{UA}}$ & MLI
& WA & F1 & $\mathrm{MEI}_{\mathrm{UA}}$ & MLI \\

\midrule

\multirow{2}{*}{GCNet}
& SMR 
&  $68.07_{\pm3.55}$&  $68.57_{\pm3.43}$&  $38.45_{\pm16.94}$&  $13.20_{\pm3.21}$&  $64.88_{\pm3.01}$&  $65.17_{\pm3.30}$&  $52.18_{\pm34.67}$&  $15.87_{\pm4.57}$&  $64.56_{\pm3.08}$&  $65.11_{\pm2.70}$&  $41.05_{\pm32.08}$&  $18.19_{\pm5.21}$
  \\

& IMR 
&  $68.99_{\pm3.54}$&  $69.35_{\pm3.47}$&  $32.71_{\pm14.74}$&  $11.27_{\pm3.69}$&  $66.24_{\pm1.79}$&  $66.69_{\pm1.56}$&  $64.24_{\pm19.16}$&  $12.15_{\pm3.38}$&  $65.00_{\pm3.53}$&  $65.41_{\pm3.27}$&  $77.01_{\pm9.12}$&  $16.99_{\pm4.04}$
  \\

\midrule
\multirow{2}{*}{RedCore}
& SMR 
&  $60.35_{\pm1.47}$&  $60.60_{\pm1.32}$&  $10.42_{\pm6.40}$&  $15.73_{\pm1.34}$&  $54.68_{\pm0.73}$&  $54.79_{\pm0.74}$&  $3.60_{\pm4.78}$&  $22.50_{\pm1.53}$&  $48.62_{\pm2.04}$&  $47.33_{\pm3.74}$&  $2.77_{\pm3.59}$&  $28.95_{\pm3.57}$

  \\

& IMR 
&  $58.03_{\pm2.02}$&  $58.32_{\pm1.91}$&  $5.24_{\pm5.12}$&  $14.86_{\pm2.40}$&  $54.18_{\pm1.13}$&  $54.22_{\pm1.29}$&  $9.78_{\pm9.00}$&  $22.03_{\pm2.13}$&  $48.88_{\pm1.66}$&  $48.07_{\pm2.54}$&  $15.30_{\pm9.15}$&  $26.20_{\pm1.34}$

  \\

\midrule
\multirow{2}{*}{Ada2I}
& SMR 
&  $72.07_{\pm1.00}$&  $72.01_{\pm1.08}$&  $17.45_{\pm1.54}$&  $19.69_{\pm4.27}$&  $71.39_{\pm1.39}$&  $71.44_{\pm1.41}$&  $17.05_{\pm4.25}$&  $27.07_{\pm5.26}$&  $71.75_{\pm1.84}$&  $71.78_{\pm1.77}$&  $22.02_{\pm7.26}$&  $32.42_{\pm6.46}$

  \\

& IMR 
&  $71.80_{\pm1.21}$&  $71.65_{\pm1.45}$&  $19.08_{\pm3.66}$&  $16.95_{\pm3.35}$&  $72.38_{\pm0.58}$&  $72.56_{\pm0.48}$&  $20.76_{\pm1.77}$&  $23.88_{\pm3.96}$&  $67.83_{\pm3.23}$&  $67.89_{\pm3.19}$&  $14.69_{\pm7.76}$&  $24.53_{\pm2.70}$

  \\
\midrule

\multirow{2}{*}{Method} 
& \multirow{2}{*}{\textbf{Setting (AL)}} 
& \multicolumn{4}{c}{0.1 / (0.1, 0.1)} 
& \multicolumn{4}{c}{0.3 / (0.4, 0.2)} 
& \multicolumn{4}{c}{0.5 / (0.3, 0.7)} \\

\cmidrule(lr){3-6}
\cmidrule(lr){7-10}
\cmidrule(lr){11-14}

& 
& WA & F1 & $\mathrm{MEI}_{\mathrm{UA}}$ & MLI
& WA & F1 & $\mathrm{MEI}_{\mathrm{UA}}$ & MLI
& WA & F1 & $\mathrm{MEI}_{\mathrm{UA}}$ & MLI \\

\midrule

\multirow{2}{*}{GCNet}
& SMR 
&  $73.44_{\pm2.82}$&  $73.83_{\pm2.58}$&  $11.24_{\pm6.63}$&  $11.31_{\pm1.23}$&  $70.09_{\pm8.31}$&  $70.06_{\pm8.87}$&  $24.74_{\pm9.21}$&  $16.15_{\pm2.44}$&  $73.55_{\pm1.04}$&  $73.71_{\pm1.02}$&  $40.32_{\pm30.70}$&  $18.23_{\pm3.06}$
  \\

& IMR 
&  $75.34_{\pm2.28}$&  $75.51_{\pm2.25}$&  $9.51_{\pm6.44}$&  $13.22_{\pm3.37}$&  $73.84_{\pm5.17}$&  $73.97_{\pm4.88}$&  $25.97_{\pm20.36}$&  $18.00_{\pm1.37}$&  $66.48_{\pm8.18}$&  $65.68_{\pm10.43}$&  $20.88_{\pm22.84}$&  $18.13_{\pm5.53}$
  \\

\midrule
\multirow{2}{*}{RedCore}
& SMR 
&  $62.35_{\pm1.50}$&  $62.37_{\pm1.55}$&  $68.35_{\pm17.09}$&  $15.12_{\pm2.22}$&  $56.86_{\pm0.71}$&  $56.76_{\pm0.77}$&  $66.52_{\pm11.27}$&  $20.85_{\pm2.38}$&  $48.99_{\pm4.45}$&  $48.41_{\pm5.02}$&  $53.33_{\pm37.97}$&  $23.37_{\pm4.38}$

  \\

& IMR 
&  $62.71_{\pm0.71}$&  $62.75_{\pm0.69}$&  $60.44_{\pm13.37}$&  $14.76_{\pm2.15}$&  $58.63_{\pm2.55}$&  $58.46_{\pm2.81}$&  $61.94_{\pm13.93}$&  $20.40_{\pm3.49}$&  $46.58_{\pm2.85}$&  $44.85_{\pm3.76}$&  $60.46_{\pm19.69}$&  $24.21_{\pm1.79}$

  \\

\midrule
\multirow{2}{*}{Ada2I}
& SMR 
&  $79.85_{\pm1.37}$&  $80.08_{\pm1.32}$&  $6.78_{\pm5.18}$&  $13.94_{\pm1.75}$&  $75.95_{\pm3.15}$&  $76.15_{\pm3.18}$&  $7.98_{\pm6.37}$&  $18.13_{\pm1.79}$&  $75.21_{\pm0.71}$&  $75.41_{\pm0.59}$&  $12.41_{\pm6.55}$&  $25.63_{\pm4.16}$

  \\

& IMR 
&  $79.05_{\pm2.07}$&  $79.31_{\pm1.97}$&  $8.75_{\pm6.92}$&  $13.17_{\pm1.79}$&  $75.62_{\pm5.14}$&  $75.83_{\pm5.17}$&  $5.09_{\pm3.85}$&  $20.02_{\pm2.65}$&  $73.02_{\pm5.95}$&  $73.24_{\pm5.96}$&  $6.58_{\pm4.48}$&  $28.12_{\pm4.95}$

  \\
\midrule

\multirow{2}{*}{Method} 
& \multirow{2}{*}{\textbf{Setting  (VL)}} 
& \multicolumn{4}{c}{0.1 / (0.1, 0.1)} 
& \multicolumn{4}{c}{0.3 / (0.1, 0.5)} 
& \multicolumn{4}{c}{0.5 / (0.2, 0.8)} \\

\cmidrule(lr){3-6}
\cmidrule(lr){7-10}
\cmidrule(lr){11-14}

& 
& WA & F1 & $\mathrm{MEI}_{\mathrm{UA}}$ & MLI
& WA & F1 & $\mathrm{MEI}_{\mathrm{UA}}$ & MLI
& WA & F1 & $\mathrm{MEI}_{\mathrm{UA}}$ & MLI \\

\midrule

\multirow{2}{*}{GCNet}
& SMR 
&  $77.82_{\pm0.81}$&  $77.75_{\pm0.75}$&  $43.86_{\pm19.16}$&  $11.00_{\pm1.87}$&  $77.94_{\pm1.61}$&  $77.86_{\pm1.53}$&  $75.00_{\pm17.23}$&  $17.20_{\pm3.21}$&  $75.04_{\pm2.19}$&  $74.77_{\pm2.08}$&  $80.74_{\pm12.41}$&  $20.08_{\pm2.67}$
 \\

& IMR 
&  $77.97_{\pm1.37}$&  $77.75_{\pm1.59}$&  $61.14_{\pm23.80}$&  $13.20_{\pm1.88}$&  $72.51_{\pm5.51}$&  $71.68_{\pm5.68}$&  $43.98_{\pm26.49}$&  $14.96_{\pm2.87}$&  $67.69_{\pm4.01}$&  $67.10_{\pm4.02}$&  $53.37_{\pm31.76}$&  $23.54_{\pm4.26}$
  \\

\midrule
\multirow{2}{*}{RedCore}
& SMR 
&  $65.62_{\pm0.53}$&  $65.70_{\pm0.98}$&  $51.91_{\pm11.02}$&  $13.80_{\pm1.23}$&  $57.50_{\pm1.91}$&  $56.75_{\pm2.63}$&  $36.30_{\pm14.96}$&  $22.57_{\pm3.48}$&  $53.36_{\pm1.89}$&  $52.69_{\pm2.06}$&  $41.69_{\pm13.55}$&  $29.94_{\pm2.66}$

  \\

& IMR 
&  $65.62_{\pm1.22}$&  $65.64_{\pm1.32}$&  $43.47_{\pm17.53}$&  $13.85_{\pm1.49}$&  $57.37_{\pm1.24}$&  $57.16_{\pm1.14}$&  $42.26_{\pm23.39}$&  $23.48_{\pm2.42}$&  $50.07_{\pm0.94}$&  $47.05_{\pm3.15}$&  $23.98_{\pm12.20}$&  $29.57_{\pm3.66}$

  \\

\midrule
\multirow{2}{*}{Ada2I}
& SMR 
&  $77.55_{\pm2.75}$&  $77.39_{\pm2.83}$&  $28.55_{\pm14.56}$&  $12.85_{\pm1.46}$&  $76.36_{\pm2.15}$&  $76.21_{\pm2.23}$&  $40.57_{\pm28.42}$&  $19.88_{\pm2.47}$&  $73.30_{\pm1.04}$&  $73.18_{\pm1.12}$&  $39.72_{\pm11.41}$&  $22.49_{\pm3.72}$

  \\

& IMR 
&  $77.76_{\pm2.56}$&  $77.60_{\pm2.70}$&  $42.70_{\pm32.25}$&  $14.30_{\pm1.75}$&  $73.60_{\pm3.18}$&  $73.35_{\pm3.37}$&  $30.68_{\pm16.59}$&  $20.03_{\pm3.31}$&  $68.36_{\pm4.29}$&  $67.80_{\pm4.36}$&  $13.87_{\pm13.41}$&  $22.94_{\pm4.32}$

  \\
\bottomrule
\end{tabular}
    }
\end{table*}

\begin{figure}[t!]
    \centering
    \includegraphics[width=0.9\linewidth]{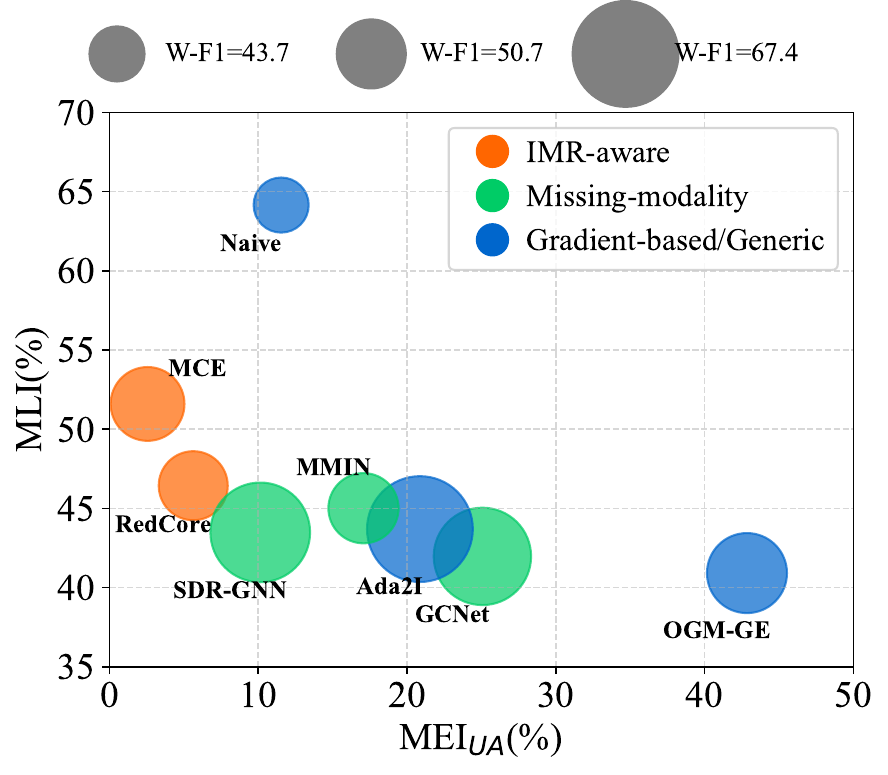}
    \caption{Performance of baselines on IEMOCAP under extreme IMR=(0.4,0.8,0.9).}
    \label{fig:extreme-imr-iemocap}
\end{figure}

\textit{Bimodal analysis under mean-matched IMR.}
Table~\ref{tab:bimodal-result} reports bimodal results under mean-matched SMR and IMR settings. 
Across all modality pairs, moving from SMR to IMR consistently increases MLI and amplifies modality imbalance, even when task-level performance changes are modest. This effect is particularly evident in settings with reduced redundancy. For example, in the AV configuration, GCNet’s $\mathrm{MEI}_{\mathrm{UA}}$ rises sharply under IMR at higher missing levels (e.g., from $41.05$ to $77.01$ at $0.5$), indicating strong dominance of the less-missing modality alongside elevated MLI.



\textit{MEI and MLI under mean-matched settings.}
Building on the SMR analysis, we examine how modality equity and learning dynamics change under mean-matched IMR (\textbf{RQ2}). Across datasets, IMR consistently worsens modality-level behavior compared to SMR at the same mean missing level. On IEMOCAP, all representative methods (GCNet, RedCore, and Ada2I) exhibit markedly higher MLI under IMR. For example, RedCore’s MLI increases from $34.8$ to $43.1$ under SMR and from $41.1$ to $55.9$ under IMR when moving from low to high missing regimes (Table~\ref{tab:mei-mli-merge}). On CMU-MOSI, GCNet and RedCore retain relatively high $\mathrm{MEI}_{\mathrm{UA}}$ under IMR, yet their MLI still rises substantially, while Ada2I shows unstable $\mathrm{MEI}_{\mathrm{UA}}$ together with steadily increasing MLI, indicating compounded imbalance in both contribution and optimization.

Figure~\ref{fig:GCNet-prob-grad} provides a mechanistic illustration of these effects using GCNet on IEMOCAP. Under SMR, modality contribution trajectories remain stable across epochs, whereas IMR induces pronounced shifts during training, with modalities experiencing lower missing rates becoming increasingly dominant. This behavior is accompanied by higher temporal gradient volatility and reduced cross-modality similarity, explaining the elevated MLI values observed under IMR. Consistent trends are also observed in the bimodal setting (Table~\ref{tab:bimodal-result}), where reduced modality redundancy further amplifies IMR-induced imbalance.


\subsection{Beyond Mean-matched: Extreme IMR and Method Trade-offs}

To address \textbf{RQ3}, we move beyond mean-matched settings and examine model behavior under an extreme IMR configuration $(r_A, r_V, r_L) = (0.4, 0.8, 0.9)$.

\textit{Extreme IMR exposes distinct equity--stability trade-offs.}
Under this setting, different method families occupy clearly separated regions in the MEI--MLI plane on IEMOCAP (Figure~\ref{fig:extreme-imr-iemocap}). IMR-aware methods achieve relatively high MEI but incur noticeably higher MLI, while missing-modality handling models cluster in an intermediate region. Gradient-based or generic baselines exhibit divergent behaviors, either reducing MLI at the expense of low MEI or maintaining moderate equity without competitive performance.

\textit{Training-time diagnostics reveal language-locking.}
Training-time analysis on IEMOCAP and CMU-MOSI shows that moving from SMR to IMR progressively shifts contribution trajectories toward the language modality. For GCNet (Figure~\ref{fig:GCNet-prob-grad}), language dominates $p_{\text{MEI}}$ and exhibits the largest temporal gradient variation, even when its missing rate is only moderately lower than the others. Similar trends are observed for RedCore and Ada2I across both datasets (Figures~\ref{fig:redcore-diagnosis-iemocap}-\ref{fig:ada2i-diagnosis-iemocap}, reported in Appendix). These patterns indicate a \emph{language-locking} failure mode under severe imbalance.

\textit{Failure modes beyond mean-matched IMR.}
Combined with the mean-matched results and the IMR sweep results reported in Tables~\ref{tab:smr-imr-iemocap-appendix}-\ref{tab:smr-imr-mosi-appendix} and Figure~\ref{fig:mei-evaluated} (\textit{Appendix}), extreme IMR reveals a distinct trade-off that does not surface under milder imbalance. Methods that appear well-behaved under SMR or mean-matched IMR can shift to markedly different regions of the MEI-MLI plane as imbalance intensifies or the dominant modality changes. These observations answer \textbf{RQ3} by exposing failure modes, such as language locking and severe gradient dominance, that remain invisible to task-level evaluation under symmetric missing rates.



\section{Conclusion}

We presented MissBench, a benchmark and framework for multimodal affective computing that standardizes shared and imbalanced missing-rate protocols on four popular datasets. By introducing the Modality Equity Index (MEI) and Modality Learning Index (MLI), MissBench complements accuracy-based evaluation with modality- and optimization-level diagnostics. Our experiments show that models can look robust under shared missing rates yet still suffer from modality inequity and gradient dominance under imbalanced conditions, even at matched mean missing ratios. MissBench thus offers a practical tool for stress-testing multimodal affective models in realistic incomplete-modality settings and motivates future methods that jointly optimize task performance, modality equity, and balanced learning dynamics.

\bibliographystyle{ACM-Reference-Format}
\bibliography{references}

\pagebreak
\appendix
\section*{Appendices}

\section{Benchmark Datasets}

This section provides an overview of the benchmark datasets used in our experiments. We summarize their key characteristics, including modality composition, task settings, and data statistics, to support reproducibility and facilitate interpretation of the experimental results reported in the main paper.

\begin{table}[h!]
    \centering
    \caption{Statistical overview of the MER and MSA datasets.}
    \label{tab:dataset-statistics}
\resizebox{0.8\linewidth}{!}{%
\begin{tabular}{ccccccc}
\toprule
\multirow{2}{*}{\textbf{Dataset}} & \multicolumn{3}{c}{\textbf{Dialogues}} & \multicolumn{3}{c}{\textbf{Utterances}} \\ \cmidrule(lr){2-4} \cmidrule(lr){5-7}
 & \multicolumn{1}{c}{train} & \multicolumn{1}{c}{valid} & test & \multicolumn{1}{c}{train} & \multicolumn{1}{c}{valid} & test \\ \midrule
\textbf{IEMOCAP} & \multicolumn{2}{c}{120} & 31 & \multicolumn{2}{c}{5810} & 1623 \\
\textbf{CMU-MOSI}  & 52    & 10    & 31   & 1284  & 229   & 686  \\
\textbf{CMU-MOSEI} & \multicolumn{1}{c}{2249} & \multicolumn{1}{c}{300} & 676 & \multicolumn{1}{c}{16326} & \multicolumn{1}{c}{1871} & 4659 \\
\textbf{CH-SIMS}   & 60    & 59    & 59   & 1,368 & 456   & 457 \\
\bottomrule
\end{tabular}%
}
\end{table}

\textbf{IEMOCAP} \cite{busso2008iemocap} is a dataset of two-person interactions where actors take part in scripted and improvised conversations to express different emotions. It contains five sessions, each divided into several utterances with categorical emotion labels. Following previous work, we use the common four-class emotion setting. Since the dataset only includes training and test sets, we further split the training data into training and validation sets with a ratio of $r$ (default 0.1).

\textbf{CMU-MOSI \cite{zadeh-mosi} \& CMU-MOSEI \cite{zadeh2018mosei}} consist of monologue and movie review video clips annotated with sentiment intensity scores between $-3$ and $3$. Aligning with the sentiment analysis setting, the datasets are trained as a regression task and evaluated using binary sentiment classification by thresholding the scores at zero, where samples with scores below 0 are labeled as negative and those above 0 as positive. The official data splits are adopted in all experiments.

\textbf{CH-SIMS} \cite{yu2020ch} is a Chinese benchmark dataset designed for multimodal sentiment analysis. In contrast to the two datasets mentioned earlier, it offers both multimodal and unimodal sentiment labels, however, we make use of only the multimodal ones. The dataset contains over 2000 utterance-level video clips spanning a wide range of content such as films, television dramas, and variety programs. Each sample is manually labeled with a continuous sentiment score ranging from $-1$ (strongly negative) to $1$ (strongly positive).

\section{Multimodal Feature Extraction}

For the \textbf{IEMOCAP} dataset, we adopt the feature extraction procedures described in \cite{zhao2021MMIN} to obtain feature representations for each modality. Specifically, acoustic features are extracted using the openSMILE toolkit \cite{Eyben2010}, resulting in 130-dimensional frame-level feature vectors. For the visual modality, we use a pretrained DenseNet \cite{huang2017densely} and extract 342-dimensional sequential representations from video frames. Finally, we employ a pretrained BERT-large model \cite{devlin2019} to derive 1024-dimensional contextual word embeddings for the textual features.

For the \textbf{CMU-MOSI} and \textbf{CMU-MOSEI} datasets, we encode the textual modality with a pretrained BERT model \cite{devlin2019} and generate 768-dimensional contextual word embeddings. For the visual part, we process each video frame with the Facet toolkit \cite{imotions2017facet} and capture 35 facial action unit features. Finally, we extract acoustic features via the COVAREP toolkit \cite{Degottex2014} and obtain 74-dimensional feature vectors.

For the \textbf{CH-SIMS} dataset, we use a pre-trained Chinese BERT-base \cite{devlin2019} model to obtain 768-dimensional word vectors. For the visual modality, we employ the MTCNN \cite{zhang2016joint} algorithm together with the OpenFace2.0 toolkit \cite{Baltrusaitis2018} to perform face alignment and extract 709-dimensional frame-level features. Finally, we utilize the LibROSA \cite{McFee2015} toolkit and extract 33-dimensional acoustic features, including log F0, MFCCs, and CQT.

\section{Baseline Methods}

We categorize baseline methods into three groups according to how they address missing or imbalanced modalities. This organization clarifies their design assumptions and facilitates a structured comparison under SMR and IMR settings.

\paragraph{Missing Modality Handling Methods.}
This category includes methods explicitly designed to operate when one or more modalities are missing at inference time. These approaches typically rely on modality reconstruction, adaptive fusion, or robustness-oriented architectures to mitigate performance degradation caused by incomplete inputs.

\textbf{MMIN} \cite{zhao2021MMIN} adopts a masking-based training strategy that randomly drops modalities and learns modality-invariant representations, so that the model can maintain performance when certain modalities are absent at test time.

\textbf{GCNet}~\cite{lian2023gcnet} captures speaker and temporal dependencies with graph neural networks to handle incomplete conversations, and adopts a dual-task framework that jointly predicts target labels and reconstructs missing features to enhance robustness to modality absence.

\textbf{SDR-GNN}~\cite{fu2024sdr} integrates spectral analysis into a hypergraph framework to impute missing data while explicitly retaining high-frequency signals that are typically attenuated in conventional GNNs, thereby improving robustness under incomplete multimodal observations.

\paragraph{IMR-aware Methods}
IMR-aware methods explicitly account for imbalanced modality exposure during training. Instead of assuming uniform or random missing patterns, they introduce mechanisms to promote modality equity under asymmetric missing rates.

\textbf{RedCore}~\cite{sun2024redcore} employs variational encoders to construct robust cross-modal representations and dynamically regulates auxiliary supervision according to reconstruction difficulty, encouraging the model to focus more on hard-to-reconstruct modalities and samples.

\textbf{MCE}~\cite{zhao2025mce} optimizes training dynamics via game-theoretic evaluations and promotes semantic robustness through subset prediction, thereby encouraging balanced feature capability even under highly imbalanced missing rates.

\paragraph{Gradient-Based and Other Approaches}
This group consists of methods that do not directly model missingness or imbalance, but instead rely on optimization-level techniques or generic multimodal fusion strategies.

\textbf{OGM-GE}~\cite{peng2022balanced} introduces orthogonal gradient modulation for pairs of modalities to reduce gradient conflict and enforces gradient equilibrium across modalities, leading to more balanced and stable multimodal optimization.

\textbf{Ada2I}~\cite{nguyen2024ada2i} rectifies learning imbalances by dynamically re-weighting feature and modality contributions under the supervision of a learning-discrepancy metric, so that modalities with under-optimized features receive stronger updates.

In addition, we include a \textbf{naive missing-modality baseline} that performs simple zero imputation for absent modalities, feeding zero vectors into the fusion network whenever a modality is missing. This baseline reflects the common practice of treating missing inputs as all-zero features and serves as a lower bound for more sophisticated missing-modality handling methods.
\section{Implementation Details}

\subsection{Benchmark Setup}

\subsubsection{Unified Configuration}

The entry point of our pipeline is managed by a \texttt{ConfigManager} and a \texttt{BaseConfig} class. This structure allows for dynamic loading of model-specific configurations while maintaining a consistent interface for experimental arguments such as missing rates and hyperparameters.

To initiate an experiment, the system parses command-line arguments to select the base model and override default parameters:
\begin{minipage}{\linewidth}

\begin{lstlisting}[caption={Initialization of Configuration and Arguments}]
config_manager = ConfigManager()
args = config_manager.get_args()

# Example: Setting missing rates (Modal MR) for Audio, Visual, Language
args.modal_MR = [0.5, 0.3, 0.0]
\end{lstlisting}
\end{minipage}

The \texttt{args} object above encapsulates crucial settings, including \texttt{modal\_MR} (missing rates for A/V/L modalities), \texttt{dataset\_mode} (Conversational or Non-conversational) and other optimization hyperparameters.

\subsubsection{Data Loading Protocol}

We provide a unified \texttt{create\_dataset} interface that abstracts the complexity of different datasets (e.g., IEMOCAP, CMU-MOSI). The data loader automatically handles the generation of missing modality masks based on the specified protocols. Listing \ref{lst:dataloader} describes detail of unified data loading interface.

Inside the \texttt{Dataloader} class, specific masking protocol (SMR or IMR) is applied to simulate missing modalities. The loader returns a dictionary containing aligned features (\texttt{A\_feat}, \texttt{V\_feat}, \texttt{L\_feat}) and their corresponding availability masks.

\subsubsection{Model Construction}

To evaluate diverse architectures fairly, we wrap all models within a generic \texttt{Model} class. This wrapper dynamically instantiates the specific encoder and base model logic defined in the configuration. Beyond instantiation, it provides a standardized interface for loss computation and advanced optimization strategies.
Details can be seen in Listing \ref{lst:modelinterface}.

\subsubsection{Training and Evaluation}

The \texttt{train.py} pipeline is designed to be model-agnostic (Algorithm \ref{alg:pipeline}), incorporating two independent evaluation modules (MEI and MLI). It also standardizes the calculation of key performance metrics, including Weighted Accuracy (WA), Unweighted Accuracy (UA), and F1-score.

\begin{figure}[ht!]
\begin{minipage}{0.9\linewidth}
\centering
\begin{lstlisting}[caption={Unified Data Loading Interface}, label={lst:dataloader}]
def create_dataset(args):
    # Dynamically load dataset library (Conversational or Non-con)
    lib = find_dataset_using_name(args.dataset_mode)
    
    # Load raw data and generate missing masks
    dataset, dims = lib.load_data(
        args.data_dir, args.dataset, args.modal_MR
    )

    # Common arguments for all data loaders
    loader_args = {
        'batch_size': args.batch_size, 
        'embedding_dim': dims, 
        'regression': args.regression, 
        'batch_first': args.batch_first,
        'device': args.device
    }

    # Initialize PyTorch DataLoaders
    train_loader = lib.Dataloader(dataset['train'], **loader_args)
    valid_loader = lib.Dataloader(dataset['valid'], **loader_args)
    test_loader  = lib.Dataloader(dataset['test'],  **loader_args)

    return [train_loader, valid_loader, test_loader], dims
\end{lstlisting}
\end{minipage}
\end{figure}

\begin{figure}[ht!]
\begin{minipage}{0.9\linewidth}
\centering
\begin{lstlisting}[caption={Model Interface}, label={lst:modelinterface}]
class Model(nn.Module):
    def __init__(self, args):
        super(Model, self).__init__()
        # Dynamically find encoder and specific multimodal method
        base_encoder = find_encoder(args)
        base_model = find_basemodel(args)
        
        self.base_model = base_model(args, base_encoder)
        # Select loss function based on task type
        self.task_loss = MaskedMSELoss(args) if args.regression else MaskedCELoss(args)

    def forward(self, inputs):
        self.set_input(inputs) # Unpack targets and masks
        return self.base_model.get_outputs()

    def get_loss(self, outputs):
        # Calculate primary task loss considering the valid mask (umask)
        loss = {
            'task_loss': self.task_weight * self.task_loss(outputs[0], self.target, self.umask)
        }
        
        # Aggregate auxiliary losses from the specific base model
        sub_loss = self.base_model.get_loss(outputs)
        for k in sub_loss:
            loss[k] = sub_loss[k]
            
        return loss

    def apply_modulation(self, outputs):
        # Hook for gradient-based method (e.g., OGM, Ada2I)
        self.base_model.apply_modulation(outputs)
\end{lstlisting}
\end{minipage}
\vspace{-5em}
\end{figure}

\begin{algorithm}[t!]
\caption{Benchmark Training Pipeline}\label{alg:pipeline}
\begin{algorithmic}[1]
\State \textbf{Input:} \texttt{args}, \texttt{train\_loader}, \texttt{val\_loader}, \texttt{test\_loader}
\State \textbf{Initialize:} \texttt{model}, \texttt{optimizer}, \texttt{scheduler}
\State \textbf{Initialize Analyzers:} $\mathcal{A}_{MLI}$ (if enabled), $\mathcal{A}_{MEI}$ (if enabled)

\For{$epoch \leftarrow 1$ \textbf{to} $args.epochs$}
    \State \texttt{model.train()}
    \For{$batch$ \textbf{in} \texttt{train\_loader}}
        \State $outputs \leftarrow model(batch)$
        \State $loss \leftarrow model.get\_loss(outputs)$
        
        \If{$\mathcal{A}_{MLI}$ is enabled}
            \State $\mathcal{A}_{MLI}.step\_evaluate(model, outputs, batch)$
        \EndIf
        
        \State $loss.backward()$
        \State \texttt{model.clip\_gradients(args.max\_grad\_norm)}
        
        \If{$args.modulation$ is active}
            \State \texttt{model.apply\_modulation(outputs)}
        \EndIf
        
        \State \texttt{optimizer.step()}
    \EndFor
    
    \State $val\_metrics \leftarrow evaluate(args, model, val\_loader)$
    \State \texttt{scheduler.step()}
    
    \If{$\mathcal{A}_{MEI}$ is enabled \textbf{and} $epoch \% MEI\_epoch == 0$}
        \State $\mathcal{A}_{MEI}.evaluate(model, val\_loader)$
    \EndIf

    \If{$val\_metrics$ is best}
        \State Save checkpoint state $\theta^*$
    \EndIf
    \State Check Early Stopping
\EndFor

\State \textbf{Test Phase:}
\State Load best model state $\theta^*$
\State $test\_metrics \leftarrow \text{test}(args, model, test\_loader)$
\State \textbf{Final Analysis:}
\If{$\mathcal{A}_{MLI}$ is enabled} $\mathcal{A}_{MLI}.evaluate()$ \EndIf
\If{$\mathcal{A}_{MEI}$ is enabled} $\mathcal{A}_{MEI}.evaluate(model, test\_loader)$ \EndIf

\end{algorithmic}
\end{algorithm}

\subsection{Experimental Setup}

For all baselines, we adopt their official implementations, except for OGM-GE, for which a variant capable of handling three modalities is employed. Our benchmark is implemented in PyTorch\footnote{\url{https://pytorch.org/}} and experiment tracking is conducted using Comet.ml\footnote{\url{https://comet.ml}}. All experiments are run on a Tesla T4 GPU, and the final results are averaged over five runs with different random seeds. For consistency, all methods are trained under the same optimization budget and unified training and logging configurations across all datasets. Specifically, we use \texttt{batch\_size=16} for conversational handling methods and \texttt{batch\_size=128} for non-conversational handling methods, with \texttt{epochs=50}, \texttt{early\_stop=10}, the Adam optimizer, and a learning rate of \texttt{2e-4}.

\section{Additional Experiment Results}

\subsection{Full SMR and IMR Results on All Datasets}
We provide the complete SMR/IMR results for all benchmarks to complement the main-text tables. 
Table~\ref{tab:smr-imr-mosei-appendix} reports performance on CMU-MOSEI across different missing-rate configurations, 
while Table~\ref{tab:smr-imr-sims-appendix} presents the corresponding results on CH-SIMS.

\begin{table*}[ht!]
  \centering
  \caption{Performance comparison on CMU-MOSEI}
  \label{tab:smr-imr-mosei-appendix}
  \resizebox{0.95\linewidth}{!}{%
    \begin{tabular}{llccccccccc}
\toprule
\multirow{2}{*}{Method} &
\multirow{2}{*}{Setting} &
\multicolumn{3}{c}{0.3 / (0.1, 0.2, 0.6)} &
\multicolumn{3}{c}{0.5 / (0.2, 0.5, 0.8)} &
\multicolumn{3}{c}{0.7 / (0.4, 0.8, 0.9)} \\ 
\cmidrule(lr){3-5} \cmidrule(lr){6-8} \cmidrule(lr){9-11}
                         &     & Acc-2($\uparrow$) & MAE($\downarrow$) & Corr($\uparrow$) & Acc-2($\uparrow$) & MAE($\downarrow$) & Corr($\uparrow$) & Acc-2($\uparrow$) & MAE($\downarrow$) & Corr($\uparrow$) \\ 
\midrule
\multirow{2}{*}{MMIN}    & SMR &  $77.72_{\pm 0.47}$  &   $0.748_{\pm 0.022}$  &   $0.641_{\pm 0.014}$  &  $73.64_{\pm 2.82}$  &   $0.814_{\pm 0.015}$  &   $0.558_{\pm 0.004}$  &  $69.60_{\pm 1.42}$  &   $0.891_{\pm 0.032}$  &  $0.447_{\pm 0.041}$   \\
                         & IMR &  $71.80_{\pm 0.50}$  &   $0.855_{\pm 0.021}$  &   $0.497_{\pm 0.026}$  &  $68.97_{\pm 0.27}$  &   $0.890_{\pm 0.005}$  &   $0.427_{\pm 0.010}$  &  $68.00_{\pm 0.36}$  &   $0.916 _{\pm 0.002}$  &  $0.392_{\pm 0.011}$     \\
\midrule
\multirow{2}{*}{GCNet}   & SMR   &  $80.22_{\pm1.10}$ &  $0.706_{\pm0.024}$ &   $0.692_{\pm0.023}$&  $76.84_{\pm0.67}$ &  $0.773_{\pm0.016}$ &   $0.621_{\pm0.018}$&  $76.34_{\pm0.96}$ &  $0.781_{\pm0.014}$ &   $0.599_{\pm0.016}$   \\
                         & IMR &  $76.74_{\pm0.86}$ &  $0.776_{\pm0.016}$ &   $0.616_{\pm0.016}$&  $73.89_{\pm0.45}$ &  $0.828_{\pm0.010}$ &   $0.541_{\pm0.019}$&  $73.19_{\pm0.63}$ &  $0.839_{\pm0.013}$ &   $0.525_{\pm0.021}$     \\
\midrule
\multirow{2}{*}{RedCore} & SMR&  $77.20_{\pm0.52}$ &  $0.760_{\pm0.007}$ &   $0.631_{\pm0.009}$&  $72.57_{\pm0.79}$ &  $0.845_{\pm0.012}$ &   $0.535_{\pm0.015}$&  $68.10_{\pm0.44}$ &  $0.902_{\pm0.012}$ &   $0.443_{\pm0.014}$     \\
                         & IMR &  $70.39_{\pm0.54}$ &  $0.866_{\pm0.006}$ &   $0.491_{\pm0.010}$&  $66.45_{\pm0.53}$ &  $0.932_{\pm0.009}$ &   $0.381_{\pm0.010}$&  $65.87_{\pm0.49}$ &  $0.951_{\pm0.006}$ &   $0.352_{\pm0.018}$     \\
\midrule
\multirow{2}{*}{MCE}     & SMR &  $77.87_{\pm 0.64}$  &   $0.746_{\pm 0.020}$  &   $0.647_{\pm 0.009}$  &  $73.21_{\pm 0.76}$  &   $0.832_{\pm 0.016}$  &   $0.536_{\pm 0.020}$  &  $69.71_{\pm 1.04}$  &   $0.885_{\pm 0.024}$  &  $0.448_{\pm 0.033}$     \\
                         & IMR &  $72.25_{\pm 0.79}$  &   $0.846_{\pm 0.009}$  &   $0.513_{\pm 0.010}$  &  $67.32_{\pm 0.48}$  &   $0.920_{\pm 0.004}$  &   $0.391_{\pm 0.001}$  &  $67.73_{\pm 0.12}$  &   $0.928_{\pm 0.003}$  &  $0.373_{\pm 0.010}$     \\
\midrule
\multirow{2}{*}{Ada2I}   & SMR &  $82.56_{\pm0.39}$ &  $0.845_{\pm0.070}$ &   $0.718_{\pm0.012}$&  $80.13_{\pm0.32}$ &  $0.940_{\pm0.112}$ &   $0.680_{\pm0.004}$&  $78.36_{\pm0.47}$ &  $0.873_{\pm0.054}$ &   $0.624_{\pm0.012}$     \\
                         &IMR &  $78.96_{\pm0.44}$ &  $0.917_{\pm0.080}$ &   $0.646_{\pm0.005}$&  $75.25_{\pm0.41}$ &  $0.920_{\pm0.035}$ &   $0.573_{\pm0.012}$&  $74.51_{\pm0.91}$ &  $0.910_{\pm0.030}$ &   $0.541_{\pm0.034}$     \\
\midrule
\multirow{2}{*}{OGM-GE}  & SMR &  $82.00_{\pm 0.39}$  &   $0.683_{\pm 0.023}$  &   $0.710_{\pm 0.018}$  &  $79.69_{\pm 0.22}$  &   $0.736_{\pm 0.012}$  &   $0.660_{\pm 0.005}$  &  $76.79_{\pm 0.35}$  &   $0.790_{\pm 0.011}$  &  $0.600_{\pm 0.010}$     \\
                         & IMR &  $77.72_{\pm 0.58}$  &   $0.767_{\pm  0.006}$  &   $0.624_{\pm 0.004}$  &  $74.24_{\pm 0.37}$  &   $0.835_{\pm 0.003}$  &   $0.527_{\pm 0.004}$  &  $74.13_{\pm 0.55}$  &   $0.850_{\pm 0.021}$  &  $0.523_{\pm 0.018}$     \\ 
\bottomrule
\end{tabular}
   }
\end{table*}

\begin{table*}[ht!]
  \centering
  \caption{Performance comparison on CH-SIMS}
  \label{tab:smr-imr-sims-appendix}
  \resizebox{0.95\linewidth}{!}{%
    \begin{tabular}{llccccccccc}
\toprule
\multirow{2}{*}{Method} &
\multirow{2}{*}{Setting} &
\multicolumn{3}{c}{0.3 / (0.1, 0.2, 0.6)} &
\multicolumn{3}{c}{0.5 / (0.2, 0.5, 0.8)} &
\multicolumn{3}{c}{0.7 / (0.4, 0.8, 0.9)} \\ 
\cmidrule(lr){3-5} \cmidrule(lr){6-8} \cmidrule(lr){9-11}
                         &     & Acc-2($\uparrow$) & MAE($\downarrow$) & Corr($\uparrow$) & Acc-2($\uparrow$) & MAE($\downarrow$) & Corr($\uparrow$) & Acc-2($\uparrow$) & MAE($\downarrow$) & Corr($\uparrow$) \\ 
\midrule
\multirow{2}{*}{MMIN}    & SMR &  $78.19_{\pm 0.26}$  &   $0.480_{\pm 0.012}$  &   $0.596_{\pm 0.026}$  &  $72.21_{\pm 1.86}$  &   $0.544_{\pm 0.009}$  &   $0.485_{\pm 0.028}$  &  $67.26_{\pm 2.28}$  &   $0.594_{\pm 0.034}$  &  $0.330_{\pm 0.097}$   \\
                         & IMR &  $71.80_{\pm 0.50}$  &   $0.509_{\pm 0.010}$  &   $0.542_{\pm 0.031}$  &  $68.97_{\pm 0.27}$  &   $0.621_{\pm 0.046}$  &   $0.212_{\pm 0.190}$  &  $68.00_{\pm 0.36}$  &   $0.603_{\pm 0.013}$  &  $0.318_{\pm 0.036}$     \\
\midrule
\multirow{2}{*}{GCNet}   & SMR   &  $66.03_{\pm0.59}$ &  $0.628_{\pm0.008}$ &   $0.234_{\pm0.024}$&  $65.97_{\pm1.45}$ &  $0.645_{\pm0.010}$ &   $0.168_{\pm0.041}$&  $65.10_{\pm0.66}$ &  $0.644_{\pm0.005}$ &   $0.163_{\pm0.022}$   \\
                         & IMR &  $64.94_{\pm0.96}$ &  $0.648_{\pm0.009}$ &   $0.161_{\pm0.055}$&  $64.77_{\pm0.67}$ &  $0.648_{\pm0.002}$ &   $0.152_{\pm0.026}$&  $65.55_{\pm0.43}$ &  $0.649_{\pm0.002}$ &   $0.140_{\pm0.024}$     \\
\midrule
\multirow{2}{*}{RedCore} & SMR&  $75.25_{\pm1.57}$ &  $0.517_{\pm0.024}$ &   $0.550_{\pm0.053}$&  $70.82_{\pm1.49}$ &  $0.561_{\pm0.012}$ &   $0.445_{\pm0.023}$&  $66.95_{\pm1.39}$ &  $0.596_{\pm0.016}$ &   $0.354_{\pm0.040}$     \\
                         & IMR &  $76.28_{\pm0.63}$ &  $0.521_{\pm0.008}$ &   $0.538_{\pm0.008}$&  $66.15_{\pm1.60}$ &  $0.591_{\pm0.011}$ &   $0.355_{\pm0.037}$&  $64.34_{\pm0.12}$ &  $0.613_{\pm0.006}$ &   $0.304_{\pm0.021}$     \\
\midrule
\multirow{2}{*}{MCE}     & SMR &  $77.06_{\pm 0.81}$  &   $0.486_{\pm 0.006}$  &   $0.581_{\pm 0.027}$  &  $73.09_{\pm 0.68}$  &   $0.528_{\pm 0.013}$  &   $0.498_{\pm 0.023}$  &  $67.78_{\pm 1.83}$  &   $0.590_{\pm 0.021}$  &  $0.359_{\pm 0.033}$     \\
                         & IMR &  $74.39_{\pm 1.68}$  &   $0.514_{\pm 0.009}$  &   $0.521_{\pm 0.024}$  &  $67.44_{\pm 3.40}$  &   $0.578_{\pm 0.023}$  &   $0.372_{\pm 0.061}$  &  $68.47_{\pm 0.85}$  &   $0.582_{\pm 0.019}$  &  $0.350_{\pm 0.028}$     \\
\midrule
\multirow{2}{*}{Ada2I}   & SMR &  $61.08_{\pm3.45}$ &  $0.728_{\pm0.059}$ &   $0.179_{\pm0.062}$&  $58.55_{\pm3.35}$ &  $0.800_{\pm0.105}$ &   $0.086_{\pm0.036}$&  $59.74_{\pm4.66}$ &  $0.996_{\pm0.388}$ &   $0.084_{\pm0.087}$     \\
                         &IMR &  $60.30_{\pm1.67}$ &  $1.028_{\pm0.190}$ &   $0.111_{\pm0.019}$&  $63.31_{\pm1.43}$ &  $0.755_{\pm0.123}$ &   $0.156_{\pm0.080}$&  $58.59_{\pm3.16}$ &  $1.228_{\pm0.379}$ &   $0.026_{\pm0.058}$     \\
\midrule
\multirow{2}{*}{OGM-GE}  & SMR &  $77.16_{\pm 1.08}$  &   $0.539_{\pm 0.007}$  &   $0.506_{\pm 0.013}$  &  $72.06_{\pm 0.35}$  &   $0.572_{\pm 0.005}$  &   $0.422_{\pm 0.028}$  &  $65.67_{\pm 1.43}$  &   $0.647_{\pm 0.011}$  &  $0.191_{\pm 0.032}$     \\
                         & IMR &  $69.07_{\pm 1.38}$  &   $0.624_{\pm  0.019}$  &   $0.281_{\pm 0.065}$  &  $66.58_{\pm 1.55}$  &   $0.634_{\pm 0.020}$  &   $0.202_{\pm 0.088}$  &  $65.63_{\pm 1.90}$  &   $0.654_{\pm 0.020}$  &  $0.111_{\pm 0.104}$     \\ 
\bottomrule
\end{tabular}
   }
\end{table*}

\begin{table*}[ht!]
  \centering
  \caption{Performance comparison on IEMOCAP under different imbalanced missing rates}
  \label{tab:smr-imr-iemocap-appendix}
  \resizebox{0.95\linewidth}{!}{%
    \begin{tabular}{lcccccccccccc}
\toprule
\multirow{2}{*}{Method} &
\multicolumn{2}{c}{(0.1, 0.6, 0.2)} &
\multicolumn{2}{c}{(0.6, 0.1, 0.2)} &
\multicolumn{2}{c}{(0.2, 0.8, 0.5)} &
\multicolumn{2}{c}{(0.8, 0.2, 0.5)} &
\multicolumn{2}{c}{(0.4, 0.9, 0.8)} &
\multicolumn{2}{c}{(0.9, 0.4, 0.8)} \\

\cmidrule(lr){2-3}
\cmidrule(lr){4-5}
\cmidrule(lr){6-7}
\cmidrule(lr){8-9}
\cmidrule(lr){10-11}
\cmidrule(lr){12-13}

 & WA & F1 & WA & F1 & WA & F1 & WA & F1 & WA & F1 & WA & F1 \\
\midrule

MMIN    
& $66.04_{\pm 2.63}$ & $66.34_{\pm 2.60}$ 
& $68.95_{\pm 0.65}$ & $69.05_{\pm 0.69}$ 
& $57.31_{\pm 0.88}$ & $57.46_{\pm 0.89}$
& $60.35_{\pm 1.61}$ & $60.22_{\pm 1.26}$ 
& $52.32_{\pm 2.17}$ & $52.38_{\pm 2.09}$ 
& $53.72_{\pm 0.50}$ & $53.74_{\pm 0.46}$ \\
\midrule
GCNet   
& $74.61_{\pm 2.28}$ & $74.91_{\pm 2.04}$ 
& $77.14_{\pm 2.63}$ & $76.87_{\pm 2.94}$ 
& $68.03_{\pm 4.52}$ & $68.31_{\pm 4.45}$ 
& $76.52_{\pm 2.09}$ & $76.33_{\pm 2.12}$  
& $65.24_{\pm 0.56}$ & $65.58_{\pm 0.43}$ 
& $72.52_{\pm 3.67}$ & $72.42_{\pm 3.43}$ \\
\midrule
RedCore 
& $65.94_{\pm 0.27}$ & $66.11_{\pm 0.25}$ 
& $65.94_{\pm 0.56}$ & $66.12_{\pm 0.57}$ 
& $60.05_{\pm 0.74}$ & $60.27_{\pm 0.72}$ 
& $59.97_{\pm 2.73}$ & $60.06_{\pm 2.40}$ 
& $52.61_{\pm 1.42}$ & $52.52_{\pm 1.49}$ 
& $53.34_{\pm 2.46}$ & $53.56_{\pm 2.35}$ \\
\midrule
MCE     
& $68.73_{\pm 0.84}$ & $68.84_{\pm 0.81}$ 
& $69.54_{\pm 0.62}$ & $69.49_{\pm 0.55}$ 
& $59.54_{\pm 1.20}$ & $59.67_{\pm 1.25}$ 
& $62.47_{\pm 0.89}$ & $62.46_{\pm 0.77}$ 
& $53.93_{\pm 0.81}$ & $53.92_{\pm 0.86}$ 
& $55.43_{\pm 1.55}$ & $55.18_{\pm 1.47}$ \\
\midrule
Ada2I   
& $80.12_{\pm 1.03}$ & $80.31_{\pm 0.98}$ 
& $77.89_{\pm 0.53}$ & $78.00_{\pm 0.47}$  
& $76.92_{\pm 1.73}$ & $77.14_{\pm 1.69}$  
& $74.48_{\pm 1.22}$ & $74.36_{\pm 1.38}$ 
& $73.70_{\pm 2.12}$ & $73.62_{\pm 2.74}$ 
& $71.60_{\pm 2.68}$ & $71.32_{\pm 3.10}$ \\
\midrule
OGM-GE  
& $71.39_{\pm 5.20}$ & $71.57_{\pm 5.28}$ 
& $74.80_{\pm 1.64}$ & $74.89_{\pm 1.49}$  
& $69.83_{\pm 2.83}$ & $69.92_{\pm 2.76}$ 
& $72.81_{\pm 2.17}$ & $72.88_{\pm 1.98}$ 
& $66.04_{\pm 1.32}$ & $66.48_{\pm 1.26}$ 
& $64.22_{\pm 5.56}$ & $63.57_{\pm 6.46}$ \\

\bottomrule
\end{tabular}

   }
\end{table*}

Table~\ref{tab:smr-imr-iemocap-appendix} summarizes the performance on IEMOCAP under varying imbalanced missing rates, 
and Table~\ref{tab:smr-imr-mosi-appendix} shows the results on CMU-MOSI.
These tables give a complete view of how all baselines and our method behave under both symmetric and imbalanced missing-modality regimes on each dataset.

\begin{table}[ht!]
  \centering
  \caption{Performance comparison on CMU-MOSI under different imbalanced missing rates}
  \label{tab:smr-imr-mosi-appendix}
  \resizebox{0.9\linewidth}{!}{%
    \begin{tabular}{lcccc}
\toprule
Method & Setting & Acc-2($\uparrow$) & MAE($\downarrow$) & Corr($\uparrow$) \\
\midrule

\multirow{6}{*}{MMIN}
& (0.1,0.6,0.2) & $74.13_{\pm0.72}$ & $1.009_{\pm0.019}$ & $0.635_{\pm0.010}$ \\
& (0.6,0.1,0.2) & $73.73_{\pm0.96}$ & $1.076_{\pm0.097}$ & $0.622_{\pm0.022}$ \\
& (0.2,0.8,0.5) & $67.02_{\pm0.94}$ & $1.191_{\pm0.024}$ & $0.514_{\pm0.001}$ \\
& (0.8,0.2,0.5) & $65.65_{\pm0.31}$ & $1.210_{\pm0.028}$ & $0.493_{\pm0.006}$ \\
& (0.4,0.9,0.8) & $52.03_{\pm7.26}$ & $1.397_{\pm0.077}$ & $0.263_{\pm0.166}$ \\
& (0.9,0.4,0.8) & $52.38_{\pm7.18}$ & $1.417_{\pm0.060}$ & $0.248_{\pm0.156}$ \\

\midrule

\multirow{6}{*}{GCNet}
& (0.1,0.6,0.2) & $70.78_{\pm0.61}$ & $1.157_{\pm0.017}$ & $0.507_{\pm0.017}$ \\
& (0.6,0.1,0.2) & $70.73_{\pm1.14}$ & $1.146_{\pm0.017}$ & $0.516_{\pm0.014}$ \\
& (0.2,0.8,0.5) & $66.15_{\pm0.89}$ & $1.285_{\pm0.026}$ & $0.403_{\pm0.016}$ \\
& (0.8,0.2,0.5) & $64.99_{\pm1.24}$ & $1.293_{\pm0.028}$ & $0.386_{\pm0.017}$ \\
& (0.4,0.9,0.8) & $61.83_{\pm1.34}$ & $1.330_{\pm0.025}$ & $0.314_{\pm0.010}$ \\
& (0.9,0.4,0.8) & $60.77_{\pm1.90}$ & $1.337_{\pm0.026}$ & $0.297_{\pm0.016}$ \\

\midrule

\multirow{6}{*}{RedCore}
& (0.1,0.6,0.2) & $74.03_{\pm1.50}$ & $1.072_{\pm0.023}$ & $0.578_{\pm0.004}$ \\
& (0.6,0.1,0.2) & $74.84_{\pm1.74}$ & $1.076_{\pm0.030}$ & $0.593_{\pm0.010}$ \\
& (0.2,0.8,0.5) & $65.95_{\pm0.61}$ & $1.234_{\pm0.010}$ & $0.461_{\pm0.009}$ \\
& (0.8,0.2,0.5) & $67.02_{\pm1.18}$ & $1.209_{\pm0.011}$ & $0.480_{\pm0.007}$ \\
& (0.4,0.9,0.8) & $60.41_{\pm1.00}$ & $1.364_{\pm0.006}$ & $0.355_{\pm0.013}$ \\
& (0.9,0.4,0.8) & $60.82_{\pm2.09}$ & $1.360_{\pm0.034}$ & $0.330_{\pm0.025}$ \\

\midrule

\multirow{6}{*}{MCE}
& (0.1,0.6,0.2) & $74.54_{\pm1.40}$ & $1.049_{\pm0.039}$ & $0.613_{\pm0.013}$ \\
& (0.6,0.1,0.2) & $73.93_{\pm0.69}$ & $1.049_{\pm0.018}$ & $0.611_{\pm0.011}$ \\
& (0.2,0.8,0.5) & $66.41_{\pm1.22}$ & $1.200_{\pm0.015}$ & $0.491_{\pm0.011}$ \\
& (0.8,0.2,0.5) & $66.26_{\pm1.83}$ & $1.208_{\pm0.030}$ & $0.500_{\pm0.023}$ \\
& (0.4,0.9,0.8) & $59.60_{\pm2.07}$ & $1.324_{\pm0.010}$ & $0.383_{\pm0.006}$ \\
& (0.9,0.4,0.8) & $58.99_{\pm1.07}$ & $1.377_{\pm0.034}$ & $0.334_{\pm0.021}$ \\

\midrule

\multirow{6}{*}{Ada2I}
& (0.1,0.6,0.2) & $71.29_{\pm4.56}$ & $1.416_{\pm0.134}$ & $0.562_{\pm0.043}$ \\
& (0.6,0.1,0.2) & $73.47_{\pm3.14}$ & $1.210_{\pm0.091}$ & $0.557_{\pm0.059}$ \\
& (0.2,0.8,0.5) & $66.41_{\pm1.15}$ & $1.310_{\pm0.061}$ & $0.442_{\pm0.017}$ \\
& (0.8,0.2,0.5) & $62.24_{\pm2.61}$ & $1.341_{\pm0.032}$ & $0.382_{\pm0.017}$ \\
& (0.4,0.9,0.8) & $49.08_{\pm4.04}$ & $1.724_{\pm0.202}$ & $0.118_{\pm0.082}$ \\
& (0.9,0.4,0.8) & $55.48_{\pm5.34}$ & $1.586_{\pm0.140}$ & $0.182_{\pm0.089}$ \\

\midrule

\multirow{6}{*}{OGM-GE}
& (0.1,0.6,0.2) & $71.85_{\pm1.27}$ & $1.156_{\pm0.027}$ & $0.537_{\pm0.015}$ \\
& (0.6,0.1,0.2) & $71.74_{\pm0.38}$ & $1.151_{\pm0.022}$ & $0.531_{\pm0.015}$ \\
& (0.2,0.8,0.5) & $65.39_{\pm2.24}$ & $1.290_{\pm0.032}$ & $0.391_{\pm0.054}$ \\
& (0.8,0.2,0.5) & $64.17_{\pm1.95}$ & $1.323_{\pm0.046}$ & $0.358_{\pm0.072}$ \\
& (0.4,0.9,0.8) & $52.23_{\pm8.15}$ & $1.432_{\pm0.082}$ & $0.116_{\pm0.176}$ \\
& (0.9,0.4,0.8) & $51.42_{\pm7.23}$ & $1.440_{\pm0.069}$ & $0.125_{\pm0.175}$ \\

\bottomrule
\end{tabular}

   }
\end{table}

To better understand modality imbalance and optimization behavior, we report additional MEI/MLI statistics and training-time measurements. 
Table~\ref{tab:mei-mli-cmu-mosei} summarizes $\text{MEI}_{\text{UA}}$/MLI (\%) on CMU-MOSEI for representative methods from each family, 
quantifying how severely different approaches suffer from modality imbalance during training.

\begin{table*}[ht!]
    \centering
    \caption{$\text{MEI}_{\text{UA}}$/MLI(\%) of representatives from each method-family on CMU-MOSEI.} 
    \label{tab:mei-mli-cmu-mosei}
    \resizebox{0.8\linewidth}{!}{%
    \begin{tabular}{llcccccc}
\toprule
\multirow{2}{*}{Method} &
\multirow{2}{*}{Setting} &
\multicolumn{2}{c}{0.3 / (0.1, 0.2, 0.6)} &
\multicolumn{2}{c}{0.5 / (0.2, 0.5, 0.8)} &
\multicolumn{2}{c}{0.7 / (0.4, 0.8, 0.9)} \\ 
\cmidrule(lr){3-4} \cmidrule(lr){5-6} \cmidrule(lr){7-8}
                         &     & $\mathrm{MEI}_\mathrm{UA}$& MLI &  $\mathrm{MEI}_\mathrm{UA}$& MLI &  $\mathrm{MEI}_\mathrm{UA}$& MLI \\ 
\midrule
\multirow{2}{*}{GCNet}   & SMR &  $74.25_{\pm8.58}$ &  $33.63_{\pm2.97}$&  $82.90_{\pm7.13}$ &  $41.05_{\pm1.84}$&  $89.51_{\pm7.02}$ &  $44.12_{\pm3.99}$

 \\
                         & IMR &  $62.83_{\pm15.95}$ &  $32.86_{\pm3.65}$&  $63.82_{\pm6.30}$ &  $37.65_{\pm2.67}$&  $74.61_{\pm15.01}$ &  $39.74_{\pm2.49}$
   \\
\midrule
\multirow{2}{*}{RedCore} & SMR &  $80.19_{\pm2.73}$ &  $33.92_{\pm1.27}$&  $80.52_{\pm4.42}$ &  $38.63_{\pm2.29}$&  $83.62_{\pm0.75}$ &  $43.25_{\pm2.43}$
   \\
                         & IMR  &  $80.65_{\pm1.39}$ &  $35.00_{\pm2.10}$&  $84.36_{\pm3.72}$ &  $40.40_{\pm3.51}$&  $84.83_{\pm1.73}$ &  $42.86_{\pm4.67}$
  \\
\midrule
\multirow{2}{*}{Ada2I} & SMR &  $86.82_{\pm5.45}$ &  $15.22_{\pm1.21}$&  $80.65_{\pm10.39}$ &  $13.41_{\pm2.22}$&  $89.86_{\pm7.60}$ &  $12.79_{\pm1.48}$  \\
                         & IMR  &  $81.72_{\pm10.60}$ &  $14.91_{\pm1.53}$&  $77.97_{\pm13.79}$ &  $14.41_{\pm1.19}$&  $81.27_{\pm13.01}$ &  $12.21_{\pm0.56}$
 \\
\bottomrule
\end{tabular}%
    }
    
\end{table*}

Addionally, we report ablation about under different bimodal combinations on non missing data (Table \ref{tab:ablate-non-missing}).

\begin{table*}[ht!]
  \centering
  \caption{Performance comparison under different bimodal combinations (A: Audio, V: Visual, L: Language) on non-missing data}
  \label{tab:ablate-non-missing}
  \resizebox{0.9\linewidth}{!}{%
    \begin{tabular}{l|ccc|ccc|ccc}
\toprule
\multirow{2}{*}{Method}
& \multicolumn{3}{c}{AV}
& \multicolumn{3}{c}{AL}
& \multicolumn{3}{c}{VL} \\

\cmidrule(lr){2-4}
\cmidrule(lr){5-7}
\cmidrule(lr){8-10}

& WA & F1 & MEI$_{UA}$
& WA & F1 & MEI$_{UA}$
& WA & F1 & MEI$_{UA}$ \\

\midrule
GCNet
&  $68.61_{\pm1.92}$&  $69.19_{\pm1.56}$&  $35.05_{\pm21.01}$
&  $74.62_{\pm2.71}$&  $74.83_{\pm2.75}$&  $6.80_{\pm6.16}$
&  $78.31_{\pm1.40}$&  $78.04_{\pm1.26}$&  $62.24_{\pm20.04}$ \\

RedCore
&  $72.93_{\pm0.95}$&  $72.89_{\pm0.92}$&  $13.75_{\pm1.39}$
&  $80.61_{\pm2.18}$&  $80.84_{\pm2.09}$&  $8.62_{\pm7.84}$
&  $78.69_{\pm2.85}$&  $78.50_{\pm2.94}$&  $39.77_{\pm29.06}$ \\

Ada2I
&  $62.34_{\pm0.67}$&  $62.47_{\pm0.74}$&  $9.55_{\pm6.20}$
&  $65.51_{\pm1.49}$&  $65.53_{\pm1.57}$&  $71.70_{\pm14.28}$
&  $68.73_{\pm0.74}$&  $68.73_{\pm0.64}$&  $45.48_{\pm19.01}$ \\

\bottomrule
\end{tabular}
   }
\end{table*}




\subsection{MEI/MLI Analysis and Training Cost}

\begin{figure}[ht!]
    \centering
    \includegraphics[width=0.8\linewidth]{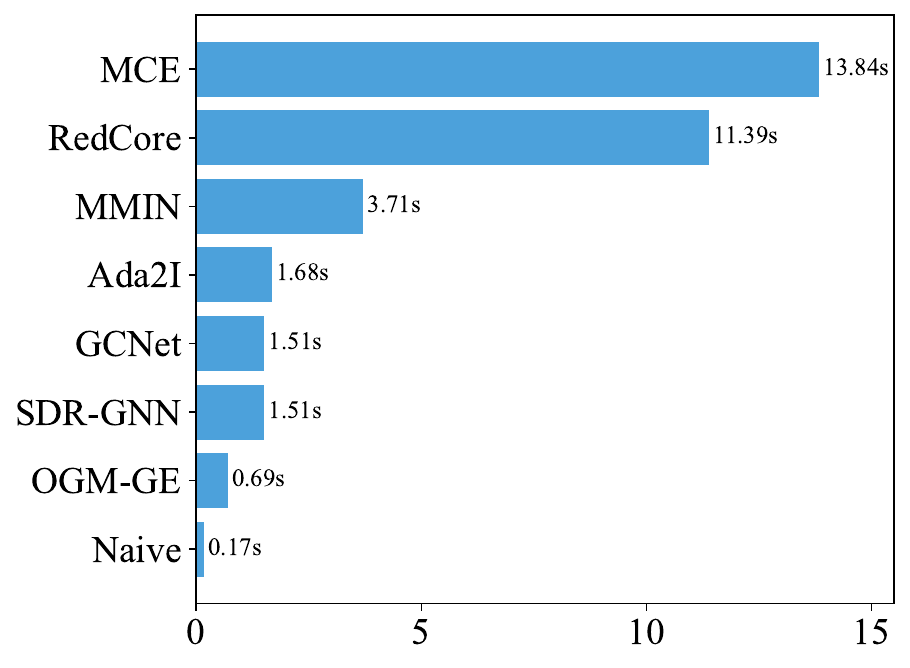}
    \caption{Estimated training time (sec/epoch) of the baselines.}
    \label{fig:training-times}
\end{figure}

\begin{figure}[ht!]
    \centering
    \includegraphics[width=\linewidth]{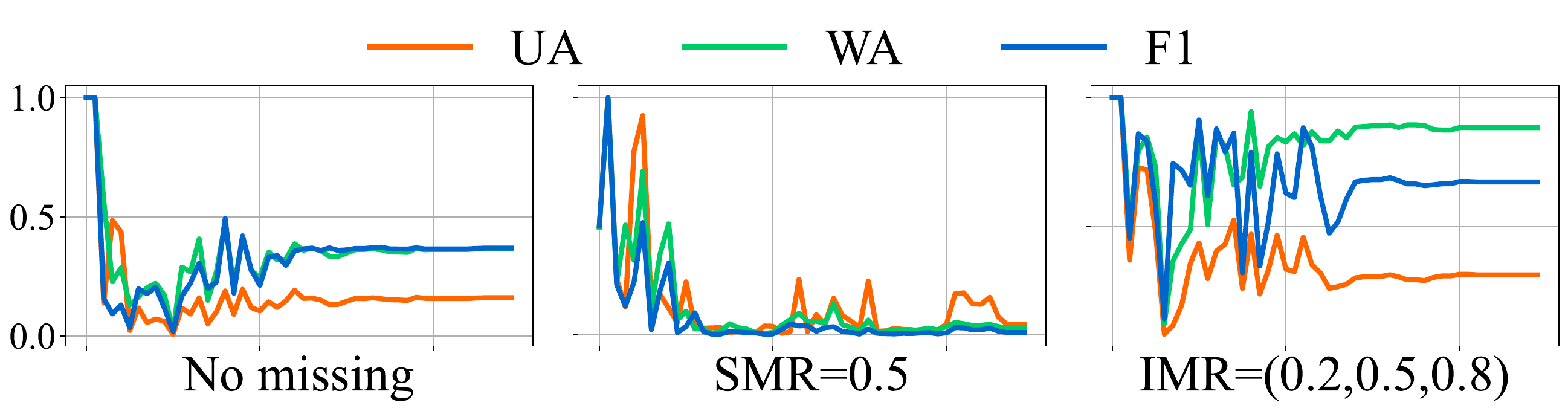}
    \caption{MEI evaluated by different task-specific metrics UA/WA/F1 during the training of GCNet on IEMOCAP.}
    \label{fig:mei-evaluated}
\end{figure}

Figure~\ref{fig:mei-evaluated} further tracks MEI over epochs for GCNet on IEMOCAP under different task-specific metrics (UA/WA/F1), 
revealing the temporal evolution of imbalance.
Finally, Figure~\ref{fig:training-times} compares the estimated training time per epoch of all baselines, 
providing a complementary view of their computational cost alongside performance.

\begin{figure}[htbp!]
    \centering
    \begin{subfigure}{\linewidth}
        \centering
        \includegraphics[width=0.85\linewidth]{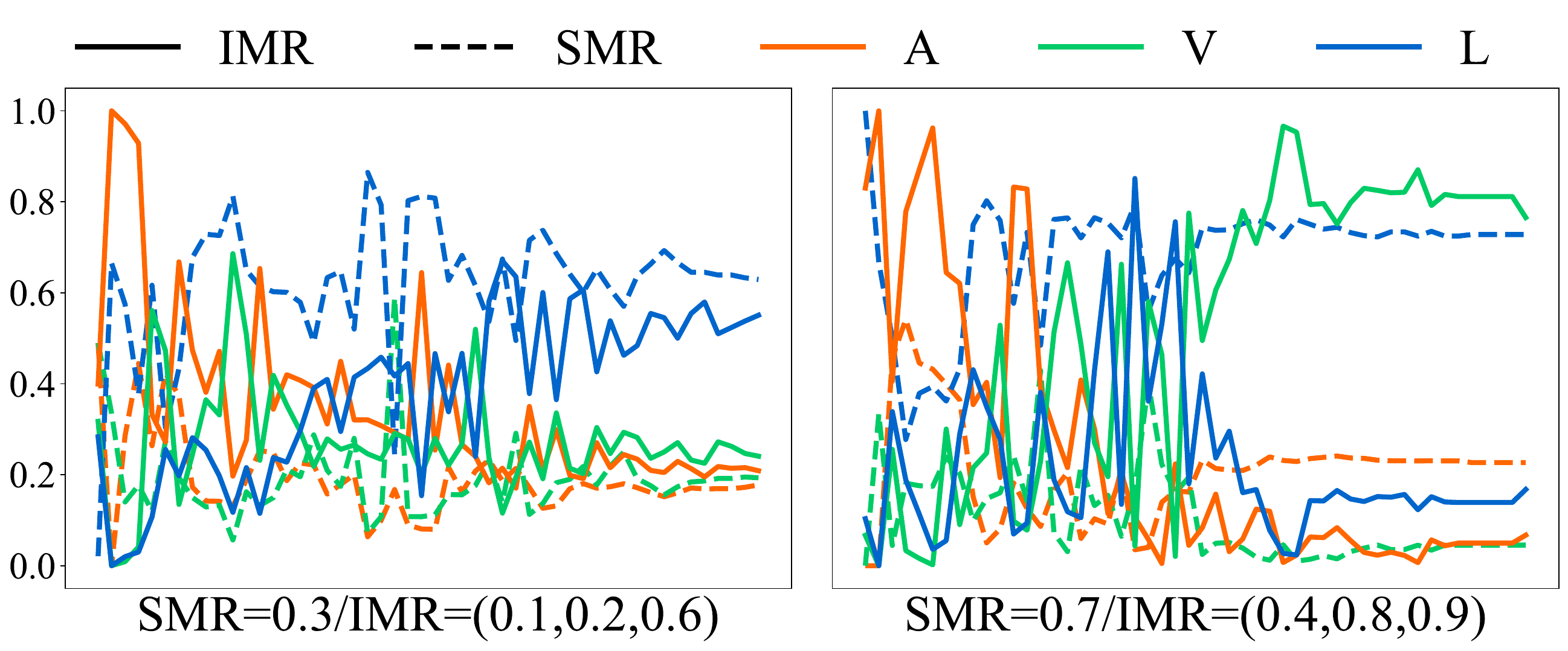}
        \subcaption{Modality contribution}
    \end{subfigure}
    
    
    \begin{subfigure}{\linewidth}
        \centering
        \includegraphics[width=0.85\linewidth]{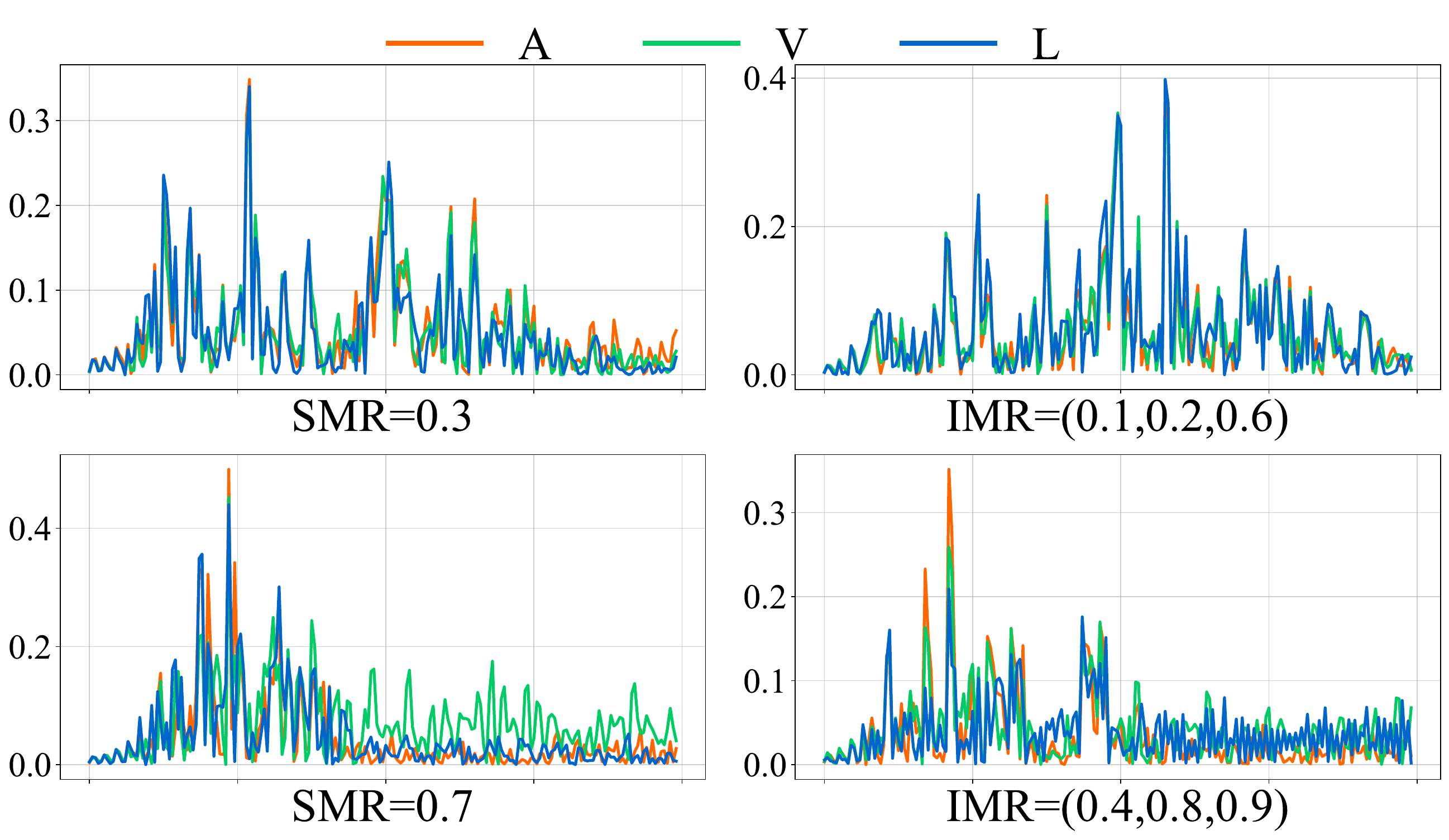}
        \subcaption{Unimodal temporal gradient variation} 
    \end{subfigure}
    \caption{Diagnosis of GCNet on a different IEMOCAP train/valid split from Figure~\ref{fig:GCNet-prob-grad}.}
    \label{fig:gcnet-diagnosis-iemocap}
\end{figure}

\pagebreak
\subsection{Case Studies: Modality Contribution and Gradient Dynamics}

We further conduct case studies to analyze how representative methods behave at the modality and optimization level. 

\begin{figure}[ht!]
    \centering
    \begin{subfigure}{\linewidth}
        \centering
        \includegraphics[width=0.8\linewidth]{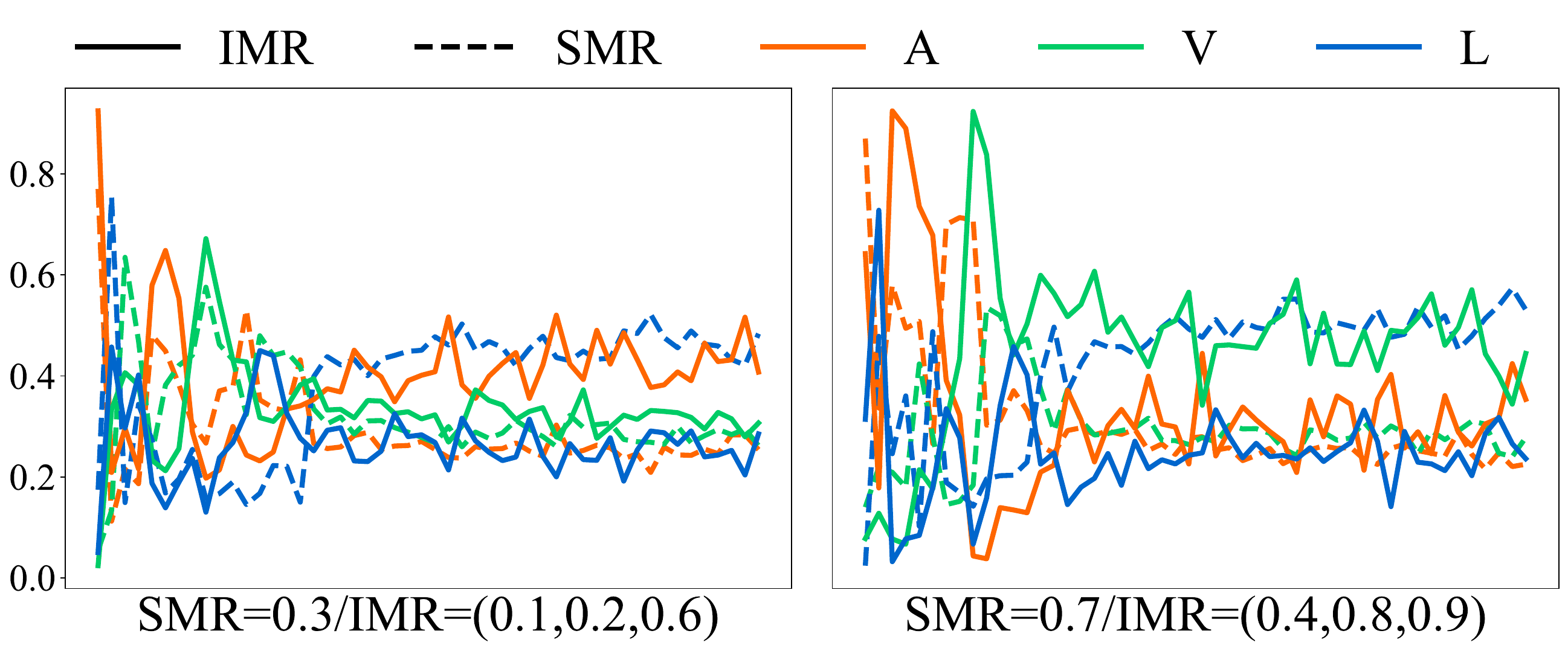}
        \subcaption{Modality contribution}
    \end{subfigure}
    \begin{subfigure}{\linewidth}
        \centering
        \includegraphics[width=0.8\linewidth]{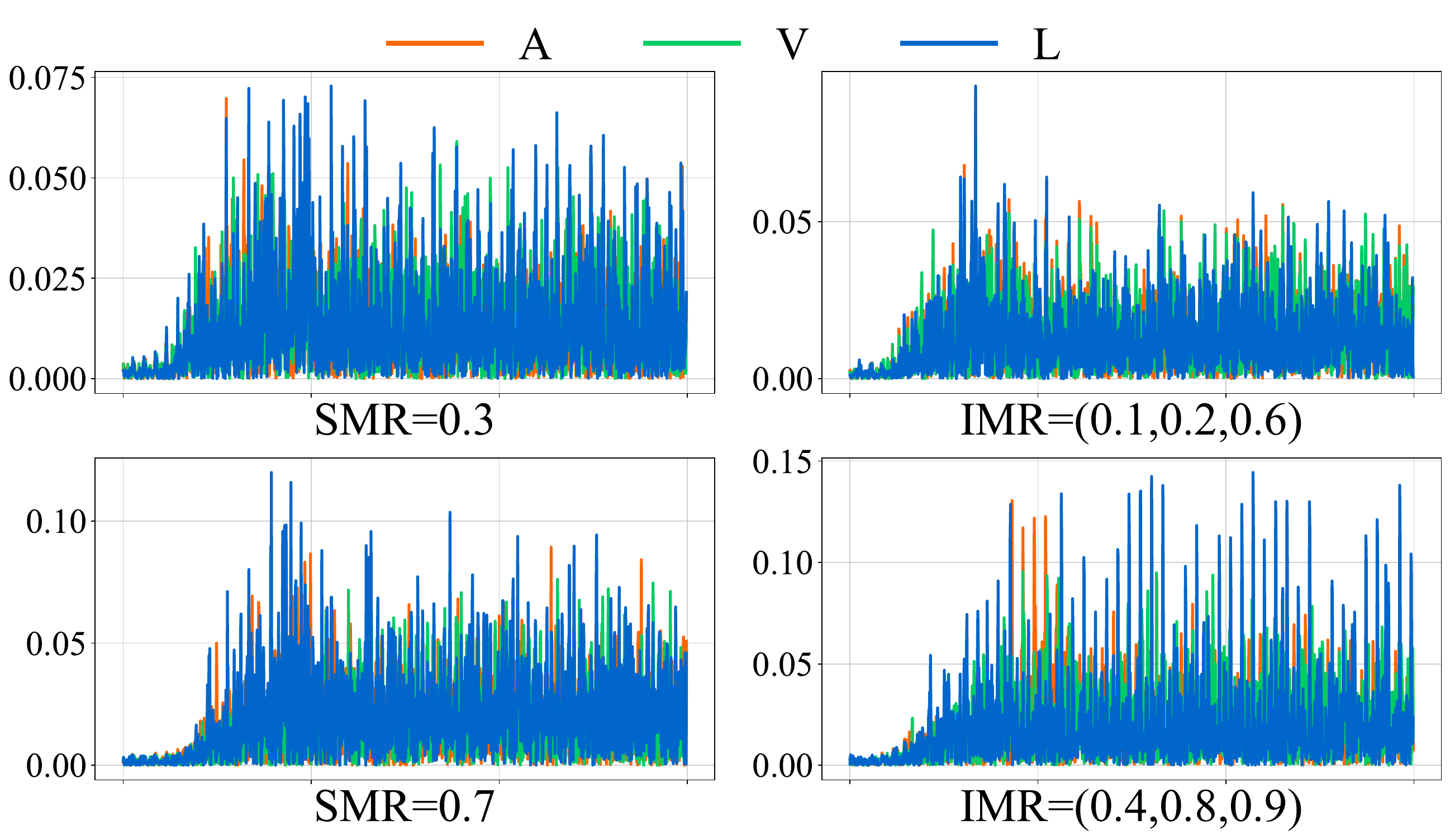}
        \subcaption{Unimodal temporal gradient variation} 
    \end{subfigure}
    \caption{Diagnosis of RedCore on IEMOCAP.}
    \label{fig:redcore-diagnosis-iemocap}
\end{figure}

\begin{figure}[ht!]
    \centering
    \begin{subfigure}{\linewidth}
        \centering
        \includegraphics[width=0.85\linewidth]{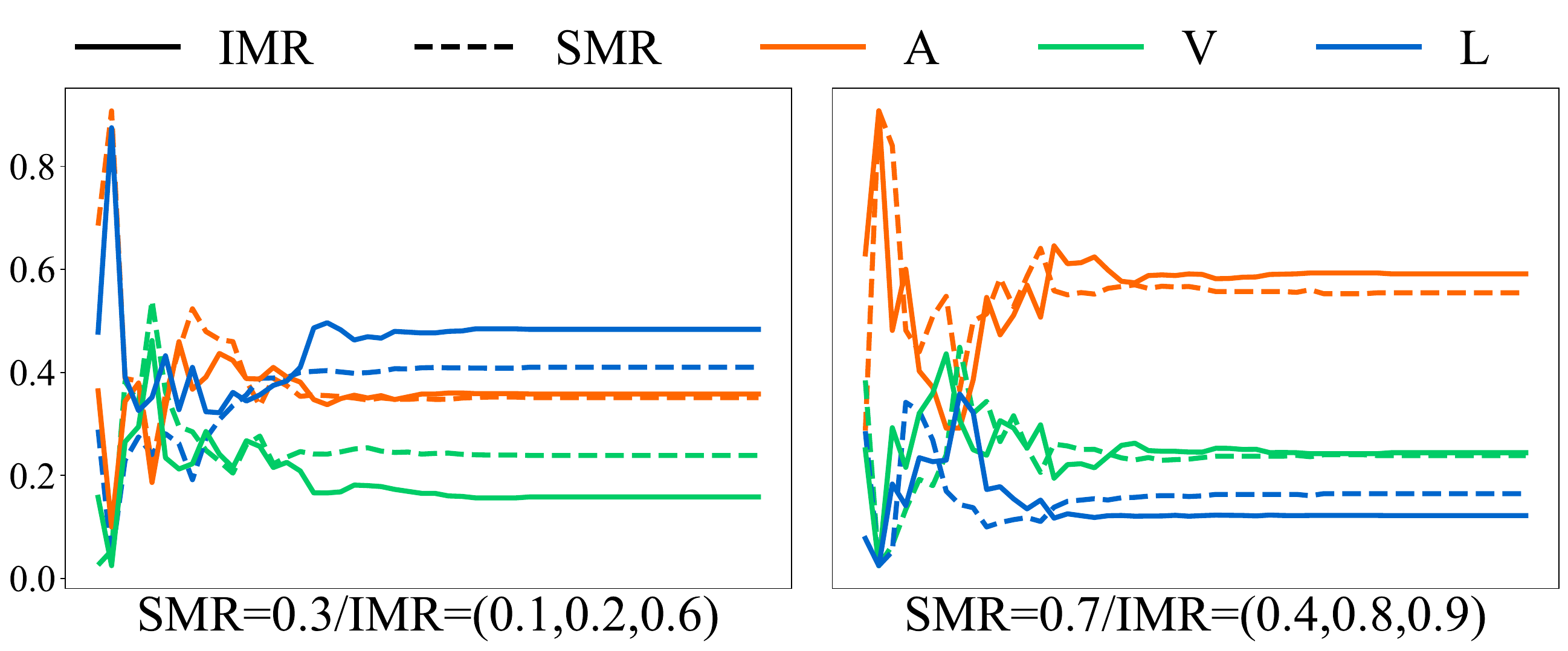}
        \subcaption{Modality contribution}
    \end{subfigure}
    \begin{subfigure}{\linewidth}
        \centering
        \includegraphics[width=0.85\linewidth]{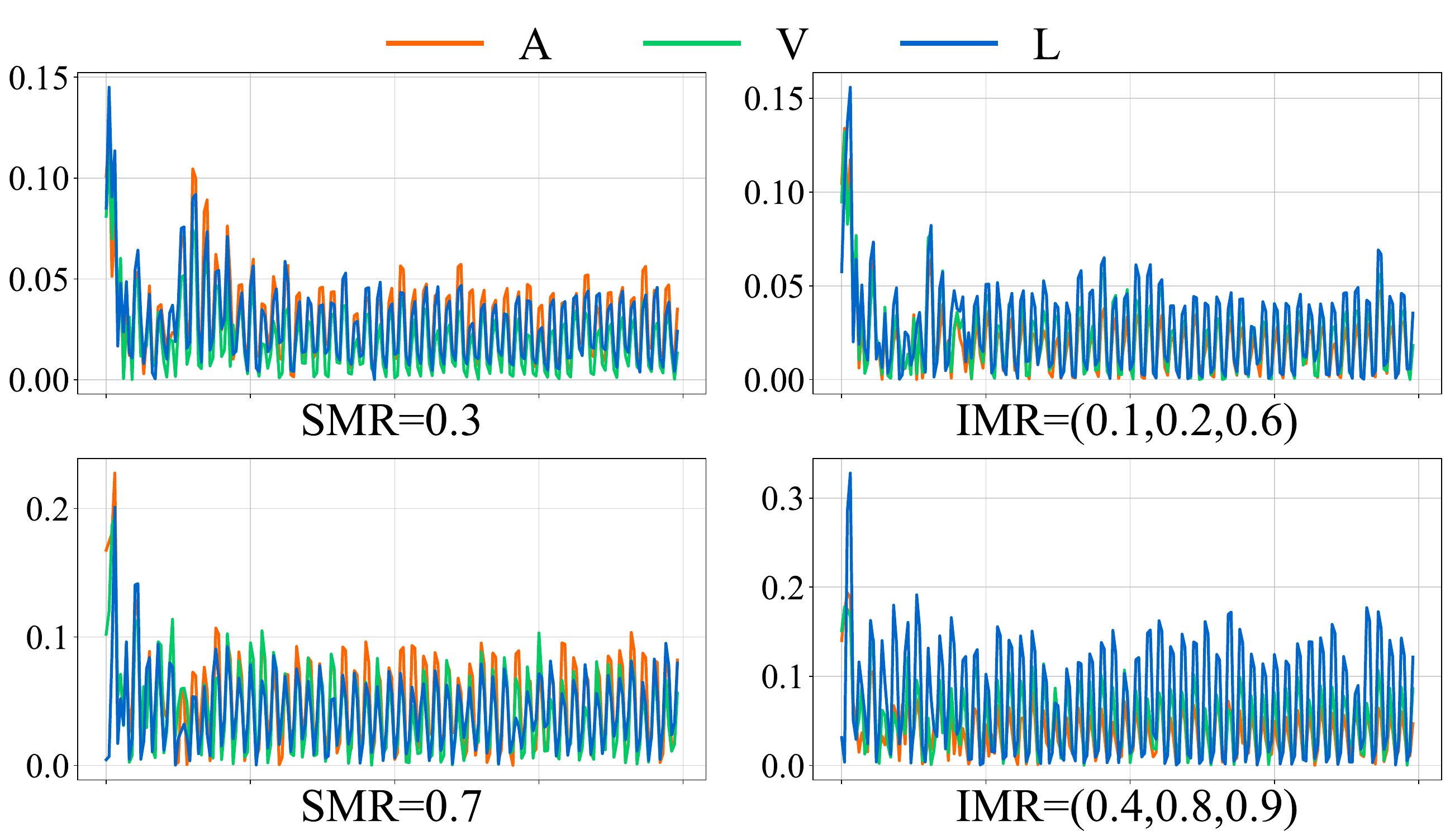}
        \subcaption{Unimodal temporal gradient variation} 
    \end{subfigure}
    \caption{Diagnosis of Ada2I on IEMOCAP.}
    \label{fig:ada2i-diagnosis-iemocap}
\end{figure}

For IEMOCAP, Figures~\ref{fig:gcnet-diagnosis-iemocap} and~\ref{fig:redcore-diagnosis-iemocap} visualize modality contribution and unimodal temporal gradient variation for GCNet and RedCore, 
while Figure~\ref{fig:ada2i-diagnosis-iemocap} presents the same diagnostics for Ada2I.

\begin{figure}[ht!]
    \centering
    \begin{subfigure}{\linewidth}
        \centering
        \includegraphics[width=0.85\linewidth]{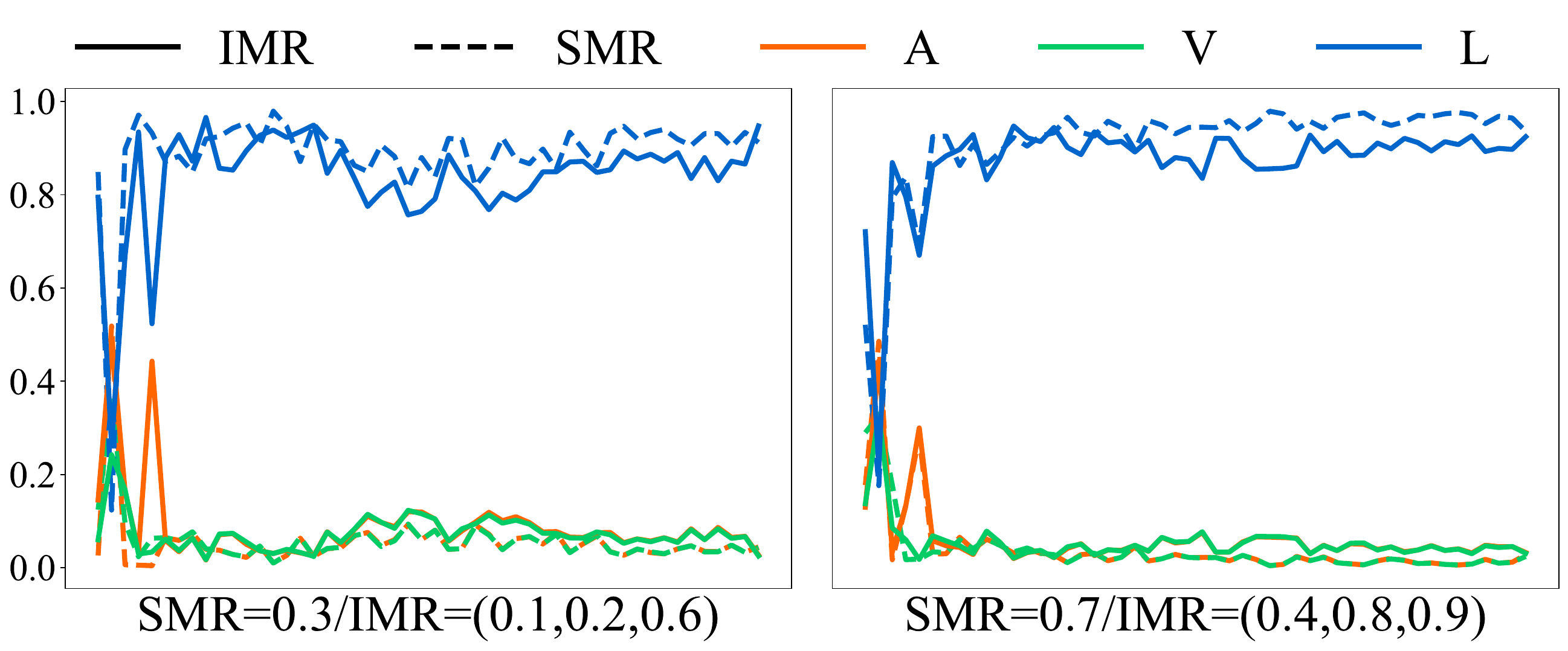}
        \subcaption{Modality contribution}
    \end{subfigure}
    \begin{subfigure}{\linewidth}
        \centering
        \includegraphics[width=0.85\linewidth]{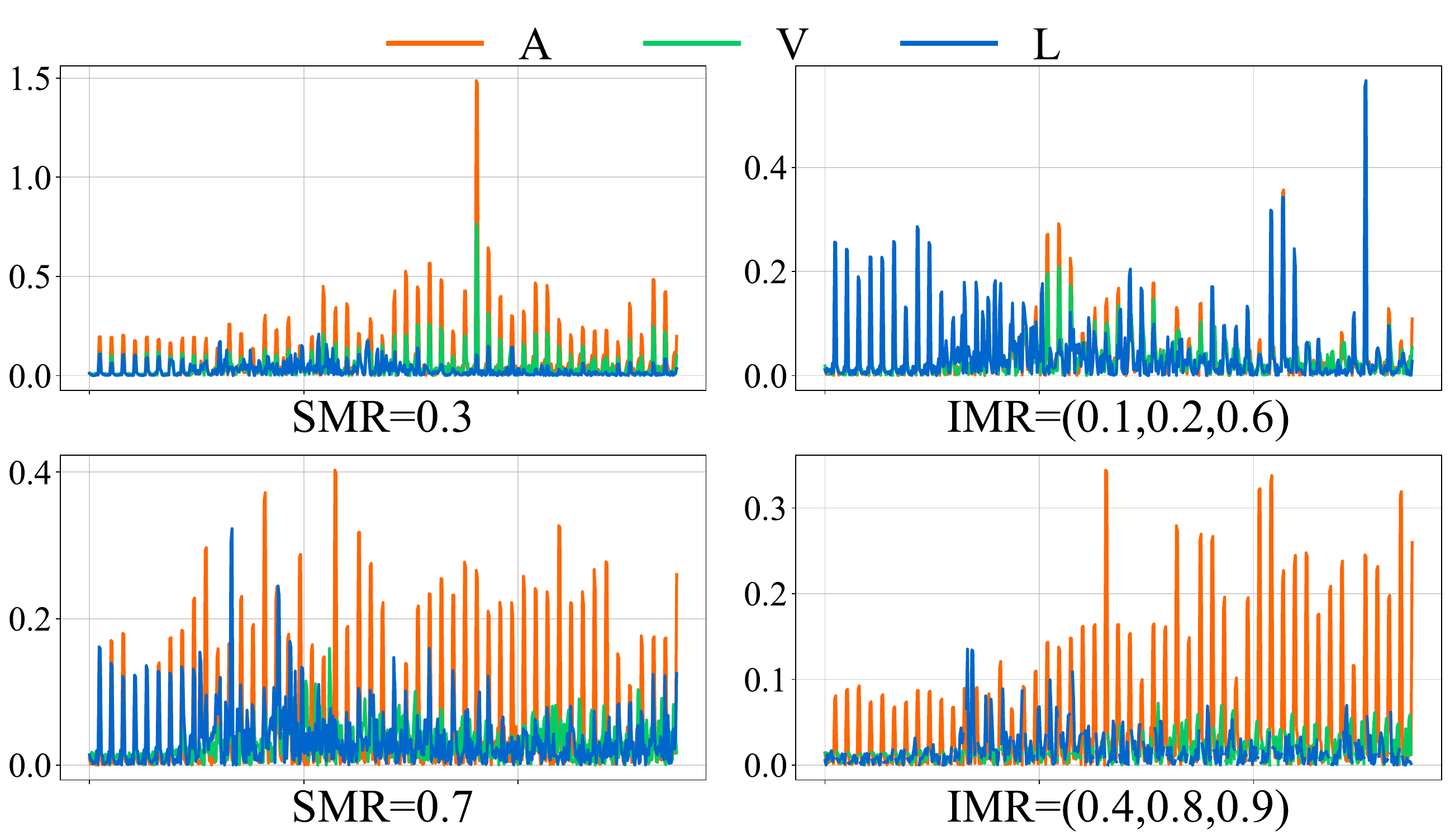}
        \subcaption{Unimodal temporal gradient variation} 
    \end{subfigure}
    \caption{Diagnosis of RedCore on CMU-MOSI.}
    \label{fig:redcore-diagnosis-mosi}
\end{figure}

\begin{figure}[ht!]
    \centering
    \begin{subfigure}{\linewidth}
        \centering
        \includegraphics[width=0.85\linewidth]{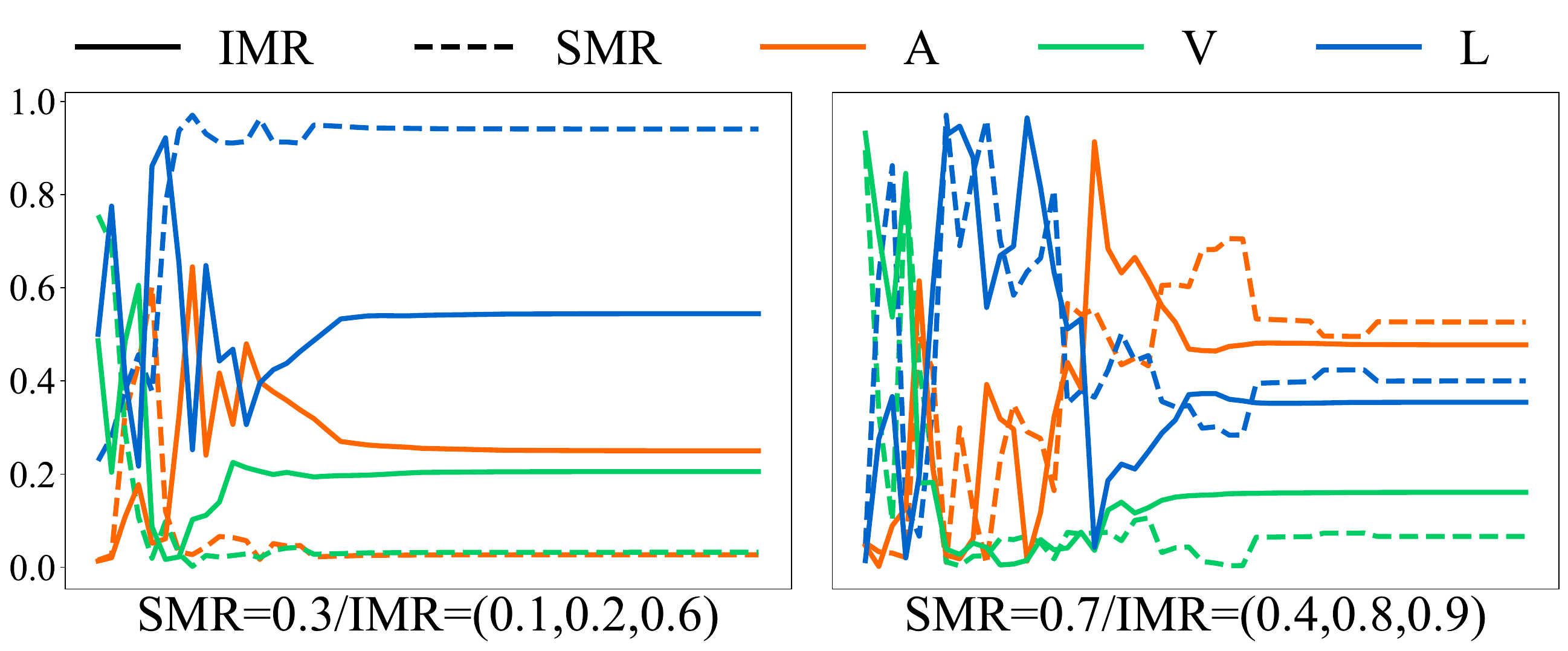}
        \subcaption{Modality contribution}
    \end{subfigure}
    \begin{subfigure}{\linewidth}
        \centering
        \includegraphics[width=0.85\linewidth]{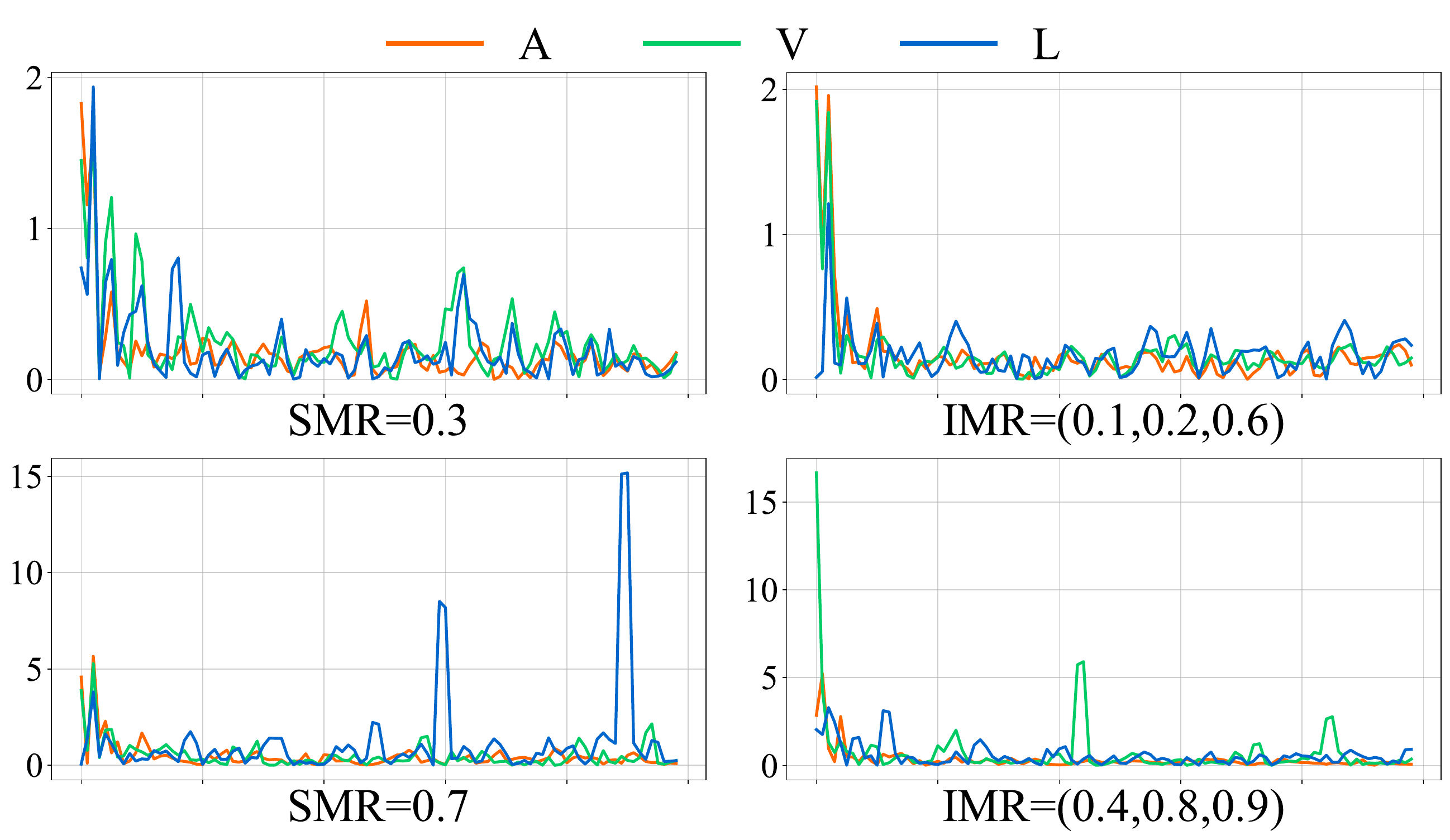}
        \subcaption{Unimodal temporal gradient variation} 
    \end{subfigure}
    \caption{Diagnosis of Ada2I on CMU-MOSI.}
    \label{fig:ada2i-diagnostic-mosi}
\end{figure}

\begin{figure}[ht!]
    \centering
    \begin{subfigure}{\linewidth}
        \centering
        \includegraphics[width=0.85\linewidth]{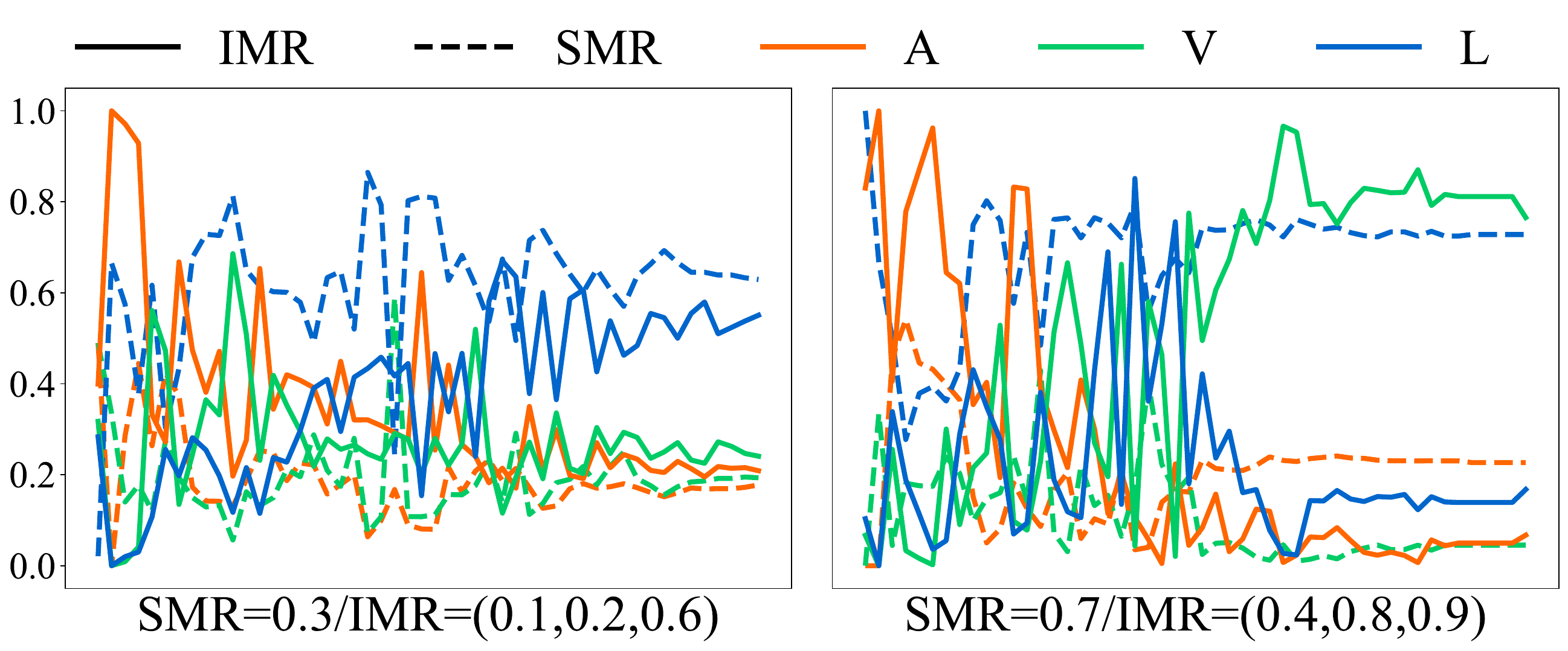}
        \subcaption{Modality contribution}
    \end{subfigure}
    \begin{subfigure}{\linewidth}
        \centering
        \includegraphics[width=0.85\linewidth]{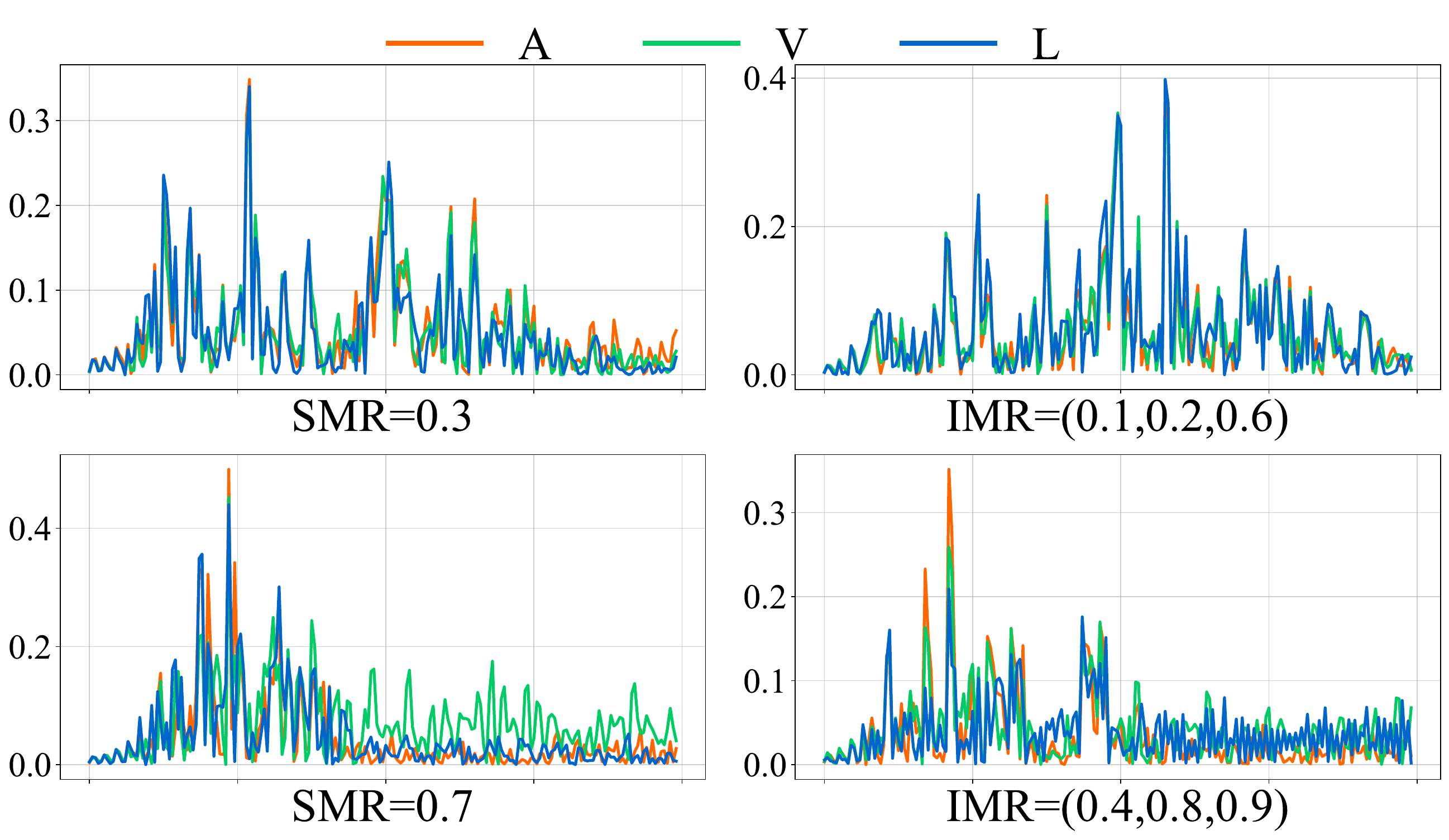}
        \subcaption{Unimodal temporal gradient variation} 
    \end{subfigure}
    \caption{Diagnosis of GCNet on CMU-MOSI.}
    \label{fig:gcnet-diagnosis-mosi}
\end{figure}

For CMU-MOSI, Figures~\ref{fig:gcnet-diagnosis-mosi}, \ref{fig:redcore-diagnosis-mosi}, and~\ref{fig:ada2i-diagnostic-mosi} provide analogous plots for GCNet, RedCore, and Ada2I, respectively.
These qualitative results complement the quantitative tables by revealing how different architectures redistribute modality contributions and stabilize (or destabilize) gradients under imbalanced missing rates. 
\clearpage

\end{document}